\documentclass[preprint,3p,english]{elsarticle}

\usepackage[T1]{fontenc}
\usepackage{lmodern}
\usepackage[utf8]{inputenc}
\usepackage{setspace}
\usepackage{booktabs}
\usepackage{tabularx}
\usepackage[table,xcdraw]{xcolor}
\usepackage{amsmath}
\usepackage{float}
\usepackage{makecell}
\usepackage{ragged2e}
\usepackage{natbib}
\usepackage{longtable}
\usepackage{ltablex}
\usepackage{array}
\usepackage{caption}
\usepackage{listings}
\usepackage{xurl}
\usepackage{hyperref,lineno,microtype,multirow}
\hypersetup{
    hidelinks,
    linkcolor={black},
    citecolor={black},
    urlcolor={black}
}

\keepXColumns    
\newcolumntype{Y}{>{\RaggedRight\arraybackslash}X}
\newcolumntype{L}[1]{>{\RaggedRight\arraybackslash}p{#1}}

\newlength{\PageFullWidthColOne}
\newlength{\PageFullWidthColTwo}
\newlength{\PageFullWidthColThree}

\newtheorem{example}{Example}

\begin{document}

\begin{frontmatter}

\title {Document Classification Pattern Recognition via Information Fusion: A Systematic Review of Multimodal and Multiview Representation Approaches\tnoteref{accepted}}

\author[opi-lis]{Marcin Micha\l{} Miro\'nczuk}\ead{marcin.mironczuk@opi.org.pl}
\address[opi-lis]{National Information Processing Institute, al. Niepodległości 188b, 00-608, Warsaw, Poland}

\cortext[corMB]{Corresponding author}
\tnotetext[accepted]{
This is the accepted manuscript version of the article published in 
\textit{Information Fusion}. The final published version is available 
at ScienceDirect via DOI: \url{https://doi.org/10.1016/j.inffus.2026.104247}.
}

\begin{abstract}
Information fusion is used widely to improve document classification by the integration of multiple data sources (multimodal) or representations (multiview). However, the field lacks a unified framework, a quantitative synthesis of its effectiveness, and clear guidance for practitioners. This systematic review addresses these gaps by analysing 139 primary studies. It introduces a formal framework to structure the field, presents the results of a qualitative analysis to identify key trends, and performs a random-effects meta-analysis (to our knowledge, the first focused on document classification) to quantify performance gains. Our meta-analysis reveals that multimodal fusion improves accuracy (mean gain of +5.28 percentage points, $p=0.0016$) significantly---the F1-score effect is directionally positive but statistically non-significant in our primary model. Multiview fusion provides consistent but modest gains for accuracy (+4.67\%), F1-score (+3.08\%), and recall (all $p<0.05$). Critically, our qualitative synthesis uncovers challenges in reproducibility in methodological rigour: only 11.8\% (multimodal) and 23.3\% (multiview) of the studies use statistical tests to validate their findings, which undermines the reliability of many of their results. This review's primary contributions are a unifying framework, the first quantitative evidence base, and data-driven guidelines. This review concludes that successful information fusion depends not on algorithmic complexity, but on the strategic alignment of the fusion method with the task context and a commitment to more rigorous validation.
\end{abstract}


\end{frontmatter}

\section{Introduction}
The exponential growth of digital information has made accurate document classification crucial across diverse domains---from legal processing to medical record management. While traditional methods have demonstrated promise, they often fail to capture the full complexity of modern documents, which frequently combine text, metadata, and multimedia elements. This limitation has driven a surge of interest in multimodal and multiview approaches that leverage information fusion to enhance classification accuracy and robustness.

Despite numerous advances in document classification techniques, four critical gaps persist in current research: (1) the absence of a unified framework for the comparison and evaluation of different information fusion strategies in document classification; (2) limited understanding of how different representation methods interact in multimodal and multiview scenarios; (3) the absence of a systematic, quantitative synthesis of evidence that compares the performance of multimodal/multiview methods against their unimodal/singleview counterparts; and (4) insufficient guidance for practitioners on the selection of appropriate fusion strategies for particular document classification tasks. This systematic review addresses these gaps by delivering a comprehensive analysis of multimodal and multiview representation approaches in document classification, with particular emphasis on information fusion strategies. 

From the perspective of information fusion, however, current work on document classification remains only loosely connected to the core theoretical frameworks developed in sensor fusion and decision making. Classical Bayesian fusion, Dempster--Shafer evidential reasoning, Kalman filtering, and probabilistic graphical models all treat fusion as the combination of uncertain evidence provided by multiple sources or sensors. By analogy, in textual settings these sources correspond to different document ``views'' (e.g., lexical features, discourse structure, citation patterns, or cross-modal signals such as images and metadata). 

Existing surveys typically adopt one of two positions. Surveys rooted in information fusion focus mainly on non-textual domains (e.g., 3D modelling, medical imaging, robotics), offering rich treatments of Bayesian and evidential fusion operators but little discussion of document representations. Conversely, multimodal and multiview Natural Language Processing (NLP) or document-classification surveys catalogue architectures and applications, yet rarely make explicit how these models instantiate formal fusion operators or propagate uncertainty across views. Consequently, to our knowledge there is no unified account of how document-level representations map onto classical fusion theories, how view-specific uncertainty and conflicts are handled, or how empirical practices in multimodal and multiview document classification relate to established information fusion paradigms.

To structure this investigation systematically, this review is guided by four primary research questions:

\begin{itemize}
    \item \textit{RQ1 (Formalization)}: How can document classification with multiview and multimodal sources be formally defined within a pattern-recognition framework, and how do existing representation and fusion mechanisms anchor to classical information fusion theories? (Sections~\ref{sec:thfund} \&~\ref{sec:inffus}) 
    
    \item \textit{RQ2 (Qualitative synthesis \& taxonomy)}: What is the state of the art in multimodal and multiview document classification, and how can this body of work be organised into a comprehensive taxonomy that identifies key research questions, discovered solutions, persistent challenges, and open problems? (Section~\ref{sec:qualanalysis})
    
    \item \textit{RQ3 (Quantitative synthesis)}: What is the synthesised magnitude, robustness, and consistency of the performance difference between multimodal/multiview and unimodal/singleview document classification methods? (Section~\ref{sec:quananalysis})
    
    \item \textit{RQ4 (Implications \& guidelines)}: Based on the formal framework developed and the results of the qualitative and quantitative analyses, what are the most promising directions for future research, and what general guidelines can be established for both practitioners and theorists in the field? (Section~\ref{sec:discussion})
\end{itemize}

Unlike broader multimodal surveys, we focus specifically on document classification and text-centric fusion, and by addressing these research questions systematically, our review makes three significant contributions to the advancement of multimodal and multiview document classification:

\begin{itemize}
    \item Proposes a comprehensive formal framework for evaluating and comparing information fusion strategies that integrates theoretical foundations with practical considerations.
    \item Introduces a novel, data-driven taxonomy that systematically organises the state of the art in multimodal and multiview classification, clarifying the interactions between different representation methods.
    \item Offers evidence-based guidelines for practitioners and researchers on the selection of fusion architectures and the identification of high-impact research topics.
\end{itemize}

The remainder of this article is organised as follows. Section~\ref{sec:relatedworks} reviews related surveys and Section~\ref{sec:systematicreviewmeth} details our systematic review methodology. The theoretical foundations are established in Sections~\ref{sec:thfund} and~\ref{sec:inffus}, covering pattern representation and information fusion as they apply to document classification. Section~\ref{sec:qualanalysis} presents a qualitative analysis of existing multimodal and multiview approaches, identifying key trends, challenges, and future research directions. This is followed by the quantitative results of the meta-analysis in Section~\ref{sec:quananalysis}. Section~\ref{sec:discussion} synthesises the review's findings to offer practical recommendations for researchers and practitioners. Section~\ref{sec:conclusion} concludes the article and summarises its key contributions.

\section{Related works}\label{sec:relatedworks}
As part of our systematic protocol, we first conducted a scoping search of the secondary literature (i.e. review articles) to confirm the novelty of our research questions. This foundational step enabled us to define the boundaries of the field, to assess the extent to which information fusion has been explored in document classification reviews, and to establish the precise gaps that a systematic review of primary studies would aim to fill. 

\subsection{Information fusion reviews analysis}\label{sec:infusrs}
Between 1998 and 2025, approximately 380 review articles~\footnote{Query time: beginning of 2025} focused on multimodal and multiview information fusion applications (see ~\ref{app:supplement} for the link to the Supplementary Materials; within the Supplementary Materials, see Section \textit{Database search queries}). This expansive growth underscores the field's evolution in computer science.

\begin{figure}[H]
\centering \includegraphics[width=1\columnwidth, height=1\textheight,keepaspectratio,trim=0 6 0 0,clip]{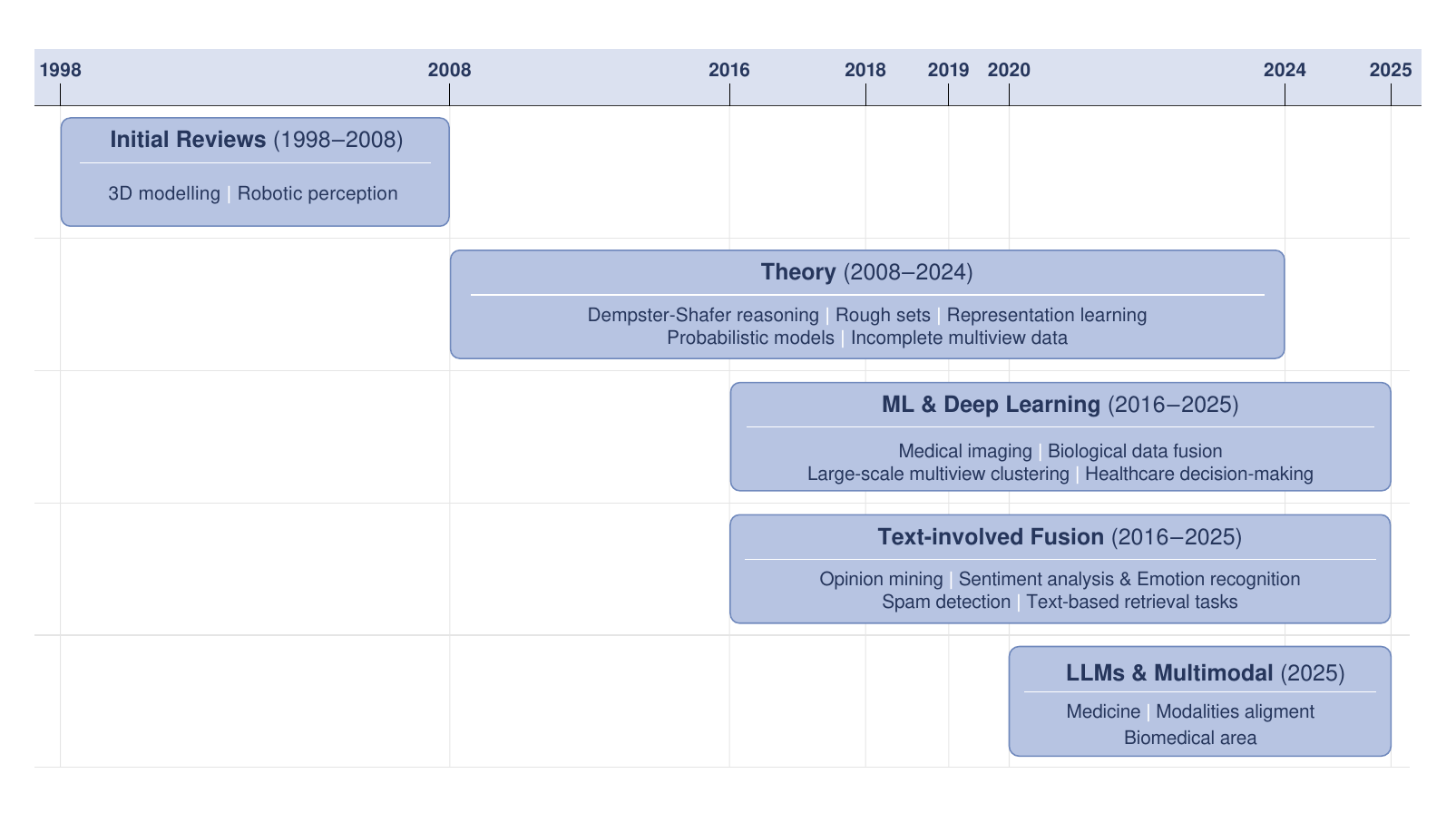}
\caption{The evolution of information fusion research through the prism of review articles.}\label{fig:informationFusionEvolution}
\end{figure}

Analysis of these reviews reveals three key trends in information fusion research (Figure~\ref{fig:informationFusionEvolution}). First, information fusion has matured primarily in nontextual domains like 3D modelling \citep{1998ardeshirThreeDimensional}, medical imaging \citep{2016duAnOverview}, and robotic perception \citep{2008zhaoSurvey,2025HuiQian}. This trend extends to biological data fusion \citep{2016liAreview} and large-scale multiview clustering techniques that integrate diverse data types, including text and images as distinct views \citep{2018wangMultiviewClust}. These works employ advanced architectures—including rational Gaussian surface fitting \citep{1998ardeshirThreeDimensional} and hierarchical B-picture coding \citep{2007sungOverviewMulti}—and often treat text as a complementary view (e.g., tags/descriptions) rather than the primary modeling target \citep{2018wangMultiviewClust}. More recent surveys highlight the growing role of multimodal fusion in medicine, in which deep-learning-based fusion techniques are being applied in oncology \citep{2025liApplication} and healthcare decision-making \citep{2025kronesReview} more frequently. Although the application of multimodal fusion for clinical predictive modelling represents a key evolution in information fusion research, its direct extension to general document classification remains largely unexplored.

Second, where text features prominently, it is often fused with other modalities in affective systems for sentiment analysis \citep{2017soleyASurveyMultimod, 2021chandraMultimodal, 2020zhangEmotion, 2019kaurMultimodalSentiment, 2024wangAreview} or spam detection \citep{2019liuOpinionSpam}. While early approaches relied on basic concatenation, more recent models---such as those in \cite{2019liuOpinionSpam}---incorporate multimodal embedded representations and probabilistic graph-based reasoning. The growing use of multimodal emotion recognition (MER) techniques, in which text, audio, and visual signals are integrated to detect cyberbullying and sentiment shifts is highlighted in \cite{2024wangAreview}. MER research utilises advanced fusion strategies that could be adapted for multimodal document classification---particularly in hierarchical fusion models that process lexical, syntactic, and semantic representations alongside nontextual data.

Third, while cutting-edge multimodal paradigms have been explored for tasks like speech emotion recognition~\citep{2021korolisDeepMultimodal}, their relevance to document classification remains underexamined. Similarly, emerging text-based retrieval tasks like person reidentification (Re-ID) \citep{2025jiangFrom} leverage text-to-image alignment, which raises new questions on how semantic text can be fused with nontextual document features.

Despite these trends, reviews cover spatial and cross-modal fusion extensively while marginalising general document classification. Exceptions, such as opinion mining \citep{2016balazasOpinionMining} and spam detection \citep{2021zhuIFSpard}, remain application-specific rather than providing broad frameworks. Notably, no work explores the fusion of intratextual multiview document representations (e.g. lexical, syntactic, and semantic layers) or cross-lingual strategies for general classification systematically. While surveys such as \citep{2018liASurvey} cover multi-view learning broadly—including some textual-only examples—they do not provide a dedicated, comprehensive review of purely textual multiview document classification.

Additional gaps persist in the handling of incomplete multiview data \citep{2024tangIncomplate}—common when text features are missing or misaligned—and in the transfer of mature methods from nontextual domains. For instance, multiscale fusion from medical imaging \citep{2016duAnOverview, 2025liApplication} could analogise document--paragraph integration, and theoretical tools like rough sets \citep{2019weiInformationFuRough}, probabilistic models \citep{2020miProbabilistic}, or Dempster--Shafer reasoning \citep{2008zhaoSurvey,2025HuiQian} might resolve conflicts across document views. Such methodological insularity also extends to multimodal text processing amid the rise of large language models (LLMs). Studies on multimodal LLMs \citep{2025songHowto} prioritise text--nontext alignments (e.g. with images) over purely textual document classification. Similarly, while reviews detail advanced fusion in the biomedical \citep{2025qianAsurvey} and medical \citep{2025xiaoAcomprehensive} domains, equivalent analyses for general document classification are largely unexplored. As LLMs expand, this underscores an urgent need—and opportunity—for robust, principled fusion techniques in text-centric tasks.

\subsection{Document classification review analysis}\label{sec:docclasssurv}
The landscape of document classification has evolved considerably, driven by the proliferation of diverse text data sources and the need for sophisticated analytical techniques. Through our review of collected articles (\ref{app:supplement} presents the exact query), we identified and categorised existing surveys according to the following key aspects: (1) \textit{Application domain}: reviews that focus on classification challenges in particular contexts, such as social media, healthcare, legal documents, and other specialised fields. Each domain presents unique requirements for text processing and analysis; (2) \textit{Task}: surveys that examine particular classification objectives, such as sentiment analysis, topic modelling, fake news detection, and other specialised tasks that require different approaches and methodologies; (3) \textit{Learning paradigms}: literature that covers various learning approaches, including supervised, semi-supervised, and self-supervised learning, and other paradigms used in document classification; (4) \textit{Data labelling}: reviews that address text annotation challenges, different label structures (binary, multilabel, hierarchical), and scenarios that involve varying amounts of labelled data; (5) \textit{Data preprocessing}: surveys that examine text preparation techniques, including tokenisation, embedding methods, and other preprocessing steps that are essential for effective classification; (6) \textit{Traditional machine learning and deep learning}: reviews that document both classical machine learning algorithms and modern deep learning approaches for document classification; (7) \textit{Model optimisation}: literature that focuses on improving model efficiency, handling long documents, and enhancing overall performance through various optimisation techniques; (8) \textit{Evaluation and validation}: surveys that pertain to evaluation metrics, benchmarking approaches, and validation methodologies for assessing classification models; and (9) \textit{Challenges and future directions}: reviews that identify ongoing challenges and emerging trends in document classification, including multilingual processing, ethical considerations, and novel methodological approaches.

\begin{figure}[H]
\centering \includegraphics[scale=0.5,keepaspectratio]{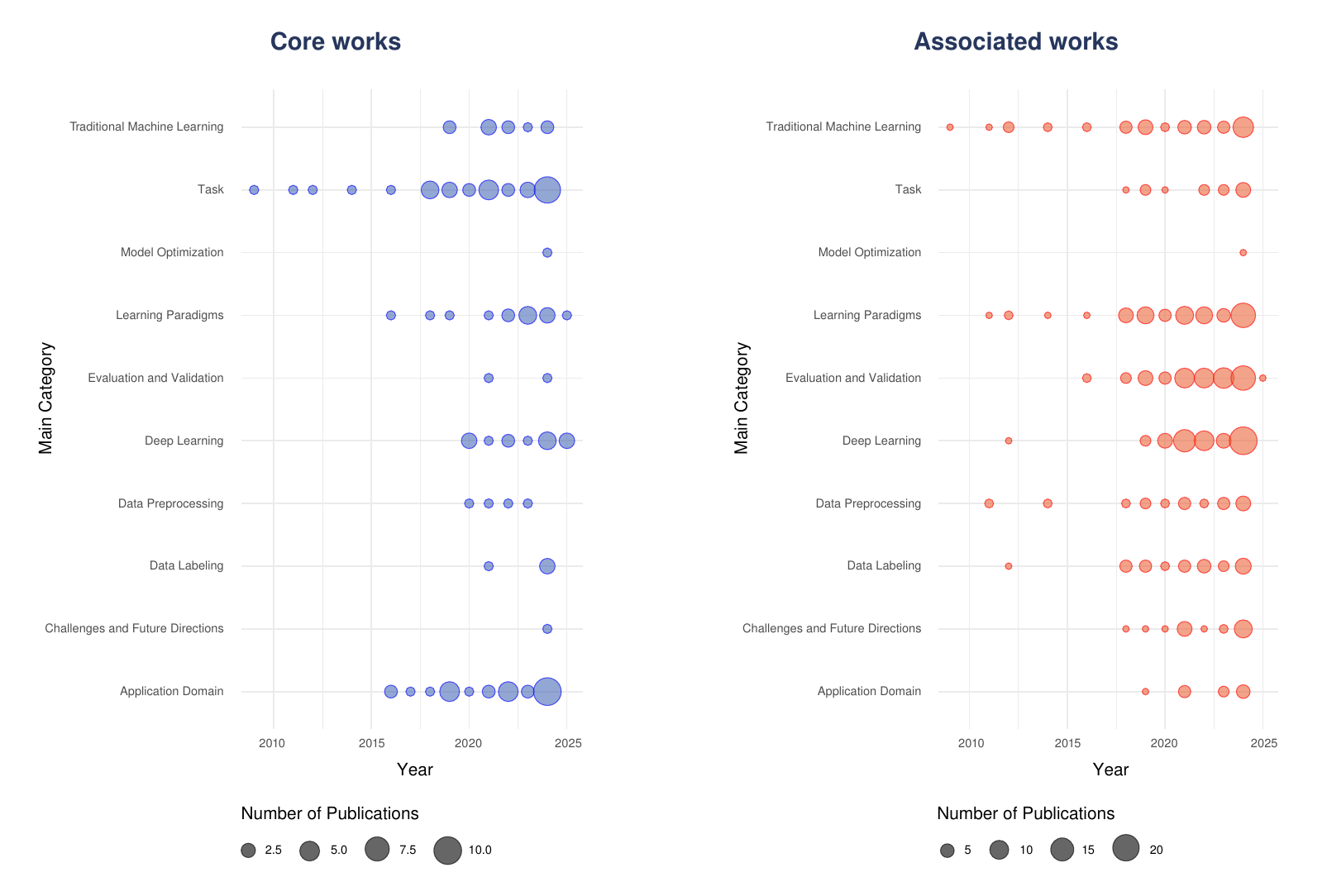}
\caption{A distribution of the studies on document classification reviews.}\label{fig:revdocclass}
\end{figure}

Figure~\ref{fig:revdocclass} presents a diverse ecosystem of studies on document classification that spans various application domains, including healthcare, social media, and language-specific classifications; tasks, including sentiment analysis and fake news detection; and learning paradigms, including supervised, semi-supervised, and transfer learning. Across these studies, the field has clearly evolved: early machine learning methods built on classical algorithms and manual feature engineering have given way to modern deep learning techniques that employ transformers, graph neural networks, and other advanced architectures. For comprehensive details, see~\ref{app:supplement} for the link to the Supplementary Materials; within the Supplementary Materials, see Section \emph{Document classification review summarisation}.

Despite this rich tapestry, a key gap has emerged in information fusion: the integration of multiple views (e.g. diverse data perspectives) and modalities (e.g. text with images or metadata). It is often found on the peripheries, appearing as auxiliary strategies to boost performance or address constraints. While related techniques like model combination (e.g. ensemble learning in which classifiers are combined, or hybrid deep learning models in which architectures are combined) have been explored widely in particular contexts (e.g. \cite{samuelak2021, simarpteetsig2024, vionadkhi2016, moludeabar2021, rini204, yonatanmam2024}), few reviews position multimodal or multiview methods as critical, standalone focuses in document classification. 

The gap is particularly evident given the promise of multimodal or multiview methods in the creation of robust, context-aware systems. Although the field is not saturated, domain-specific reviews highlight the gap: for example, fake news detection surveys identify limited text--image fusion as a critical research direction \citep{hamzaza2024}, while broader surveys note scarce integration of text with numeric data in business contexts \citep{johnfields2024}. Primary works, such as those on social media misinformation, apply multiview fusion (e.g. combining embeddings via ensembles), but limit generalisations to broader classification challenges \citep{shankarbi2023}. Similarly, sentiment analysis reviews emphasise diverse data potential without cross-domain synthesis \cite{yonatanmam2024}.

\subsection{Positioning within Information Fusion literature}\label{sec:fusion_positioning}
To rigorously position our work within the broader Information Fusion landscape, we contrast this review with foundational information fusion surveys and recent multiview and multimodal representation-learning overviews. As illustrated in Table~\ref{tab:if-survey-comparison}, existing literature can be grouped into three methodological parts:

\begin{itemize}
\item Classical Information Fusion (sensor \& uncertainty focus): Surveys such as \citep{2008zhaoSurvey,2025HuiQian,2015snidaroContextbased,2019weiInformationFuRough} provide robust mathematical grounding in evidential and uncertainty reasoning (e.g., Bayesian estimation, Dempster--Shafer theory, rough sets), and in context modelling for situation assessment. However, these frameworks focus mainly on signal-level processing and state estimation (often associated with lower JDL-Joint Directors of Laboratories-framework levels) or abstract decision-making, and do not provide the representation-learning mechanisms required to process high-dimensional unstructured data such as text documents or document layout images.

\item Aggregation \& decision theory: Works like \citep{2020miProbabilistic} focus on the mathematical properties of aggregation operators for linguistic or probabilistic-linguistic information (e.g., OWA and Bonferroni means over linguistic term sets). While methodologically rigorous, they operate fundamentally at the \textit{decision level}, assuming that information has already been encoded as decision variables, and thus sidestep the \textit{feature-level} fusion challenges inherent in document classification.

\item Multiview \& multimodal representation learning: The rise of deep learning has coincided with surveys on multiview learning \citep{2013sunAsurvey,2017zhaoMultiview,2018liASurvey} and multimodal architectures \citep{2018baltrusMultimodalMachine,2019guoDeepMultimodal,2025qinAsurvey}. These reviews excel at describing neural architectures and representation-learning objectives (e.g., deep convolutional neural networks--CNNs, recurrent neural network--RNNs, autoencoders, attention-based models), but often treat the ``fusion'' step implicitly as a learnable layer (e.g., concatenation or attention) rather than analyzing it through a formal information fusion taxonomy. They do not adopt PRISMA-style systematic-review protocols or quantitative meta-analysis to verify fusion efficacy.
\end{itemize}

\begin{table}[H]
\centering
\caption{Comparative analysis of this study against foundational Information Fusion learning surveys.}
\label{tab:if-survey-comparison}
\scriptsize
\setlength{\tabcolsep}{4pt}
\renewcommand{\arraystretch}{1.3}
\resizebox{\textwidth}{!}{%
\begin{tabular}{p{2.5cm}p{3.0cm}p{4.5cm}p{1.5cm}p{4.5cm}}
\hline
\textbf{Ref. \& Scope} & \textbf{Dominant fusion paradigm} & \textbf{Methodological focus vs.\ gap} & \textbf{Systematic protocol} & \textbf{Our added contribution relative to this work} \\
\hline

\multicolumn{5}{c}{\textbf{Part 1: Classical sensor \& context fusion}} \\
\hline
\citep{2008zhaoSurvey,2025HuiQian} \newline Multi-sensor perception & 
Bayesian, Dempster--Shafer (DS), Kalman filtering & 
\textbf{Focus:} Handling uncertainty/conflict in sensor signals and multi-sensor information fusion. \textbf{Gap:} No treatment of semantic alignment for unstructured text/image data or learned representations. & {\centering No\par} & 
Moves from signal-level uncertainty modelling to semantic-level feature integration for documents; formalises document-classification architectures as information-fusion systems. \\

\citep{2015snidaroContextbased} \newline Context/high-level information fusion & 
JDL framework & 
\textbf{Focus:} Situational awareness, context exploitation, and high-level information fusion processes. \textbf{Gap:} Does not address the feature-learning pipeline for high-dimensional unstructured document data. & {\centering No\par} & 
Applies high-level context-fusion ideas to different document ``views''
(such as structure, layout, and textual content) in document-classification pipelines. \\

\citep{2019weiInformationFuRough} \newline Rough sets & 
Granular computing, approximation spaces & 
\textbf{Focus:} Mathematical reducts, approximation operators and rule induction for multi-source information systems. \textbf{Gap:} Limited applicability to deep continuous representations (embeddings) commonly used in multimodal document models. & {\centering No\par} & 
Extends beyond rough-set-based symbolic fusion to a broader representation-learning view that encompasses both classical feature-based and deep neural representations for multimodal and multiview document fusion.
\\

\hline
\multicolumn{5}{c}{\textbf{Part 2: Aggregation operators}} \\
\hline
\citep{2020miProbabilistic} \newline Linguistic decision making & 
Aggregation operators & 
\textbf{Focus:} Mathematical properties and design of aggregation operators for probabilistic linguistic information in group decision making. \textbf{Gap:} Assumes inputs are decision variables, not raw or feature-level data streams. & {\centering No\par} & 
Extends beyond decision-level aggregation by analyzing the full pipeline that transforms raw document data into fusible feature representations and decision outputs. \\

\hline
\multicolumn{5}{c}{\textbf{Part 3: Multiview \& multimodal machine learning}} \\
\hline
\citep{2013sunAsurvey} \newline Shallow multiview & 
Canonical correlation analysis, Co-training, Co-regularization & 
\textbf{Focus:} Learning consensus and complementarity across views using mostly shallow models. \textbf{Gap:} Largely pre-deep-learning; lacks an explicit hierarchical fusion taxonomy. & {\centering No\par} & 
Frames shallow multi-view methods within a unified notation that also covers modern fusion strategies for multimodal and multiview document classification.\\

\citep{2018baltrusMultimodalMachine} \newline General multimodal & 
Early/late fusion, alignment, translation & 
\textbf{Focus:} Broad taxonomy of multimodal tasks and learning challenges across modalities. \textbf{Gap:} Descriptive rather than formal from an information-fusion perspective; no meta-analysis of fusion gains. & {\centering No\par} & 
Provides rigorous statistical evidence (meta-analysis) of fusion gains specifically for document-classification tasks and embeds these within an information-fusion framework. \\

\citep{2019guoDeepMultimodal} \newline Deep multimodal learning & 
Deep neural networks (CNNs, RNNs, attention) & 
\textbf{Focus:} Deep multimodal representation-learning architectures and application domains. \textbf{Gap:} Treats fusion mainly as an architectural component or layer; lacks an explicit information-fusion-theoretic grounding. & {\centering No\par} & 
Classifies these architectures by their fusion role and analyses their
impact on multimodal and multiview document classification. \\

\citep{2025qinAsurvey} \newline Deep multiview learning & 
Contrastive learning, graph alignment & 
\textbf{Focus:} Self-supervised and deep representation learning for multiview data. \textbf{Gap:} Emphasises representation pre-training and alignment, with limited discussion of decision-level fusion for downstream classification. & {\centering No\par} & 
Links representation alignment quality directly to document-level classification performance via meta-analysis of multimodal and multiview fusion strategies. \\

\hline
\textbf{This work} \newline \textbf{Document fusion} & 
\textbf{Formal framework to describe \textit{representation}, \textit{pattern} and \textit{model}} & 
\textbf{Contribution:} Develops a unified information-fusion framework for representations, patterns, and models, using multimodal and multiview document classification as a demonstrative lens to quantify fusion efficacy and synthesize strategies. & {\centering \textbf{Yes}\par} & 
Formally frames document classification as an information-fusion problem and supports this with a PRISMA-guided systematic review and a quantitative meta-analysis of fusion effect sizes in a unified notation. \\
\hline
\end{tabular}}
\end{table}

The comparison reveals a pivotal gap: classical information fusion \citep{2008zhaoSurvey,2015snidaroContextbased,2019weiInformationFuRough,2025HuiQian} offers rigorous treatments of uncertainty and high-level process models but lacks the semantic representation-learning capabilities required for document-level tasks, while modern deep-learning surveys on multiview and multimodal learning \citep{2013sunAsurvey,2017zhaoMultiview,2018liASurvey,2018baltrusMultimodalMachine,2019guoDeepMultimodal,2025qinAsurvey} provide rich architectural taxonomies but little explicit fusion-theoretic grounding or meta-analytic evidence. This article addresses this duality. We do not merely review document classification applications; rather, we propose a formal representation-and-pattern framework (introduced in Section~\ref{sec:thfund}) that maps the ad-hoc architectural choices of machine or deep learning (e.g., concatenation features, attention heads, concatenation layers) onto information-fusion concepts (e.g., alignment, association, and combination). Furthermore, unlike these narrative surveys, we employ a PRISMA-guided systematic approach coupled with a quantitative meta-analysis to establish the statistical significance of fusion strategies, reporting effect sizes that are not present in prior broad information fusion or multiview and multimodal-learning taxonomies.

\section{Systematic review methodology}\label{sec:systematicreviewmeth}
This section outlines our methodology for reviewing studies on multimodal and multiview document classification. We adapted key elements from the PRISMA-informed procedures \citep{Moher2010,page2021}---originally designed for medical research---to structure our analysis in the computer science domain. We systematically reviewed primary research articles that apply multimodal and multiview approaches to document classification, adhering to a structured review protocol that follows four fundamental PRISMA-aligned steps: (1) identification of relevant studies; (2) screening of titles and abstracts; (3) full-text assessment for eligibility; and (4) data extraction and analysis. This approach ensured a rigorous and reproducible synthesis of evidence on the current state of multimodal and multiview techniques for document classification.

The primary literature search was guided by the central question: \textit{What current works discuss the use of multimodal or multiview approaches in document classification?} This constitutes the core of our study and was addressed through the systematic analysis of selected articles.

\subsection{Review method: general schema}\label{sec:ReviewMethodGeneralSchema}
\begin{figure}[H]
\centering \includegraphics[scale=0.5,keepaspectratio]{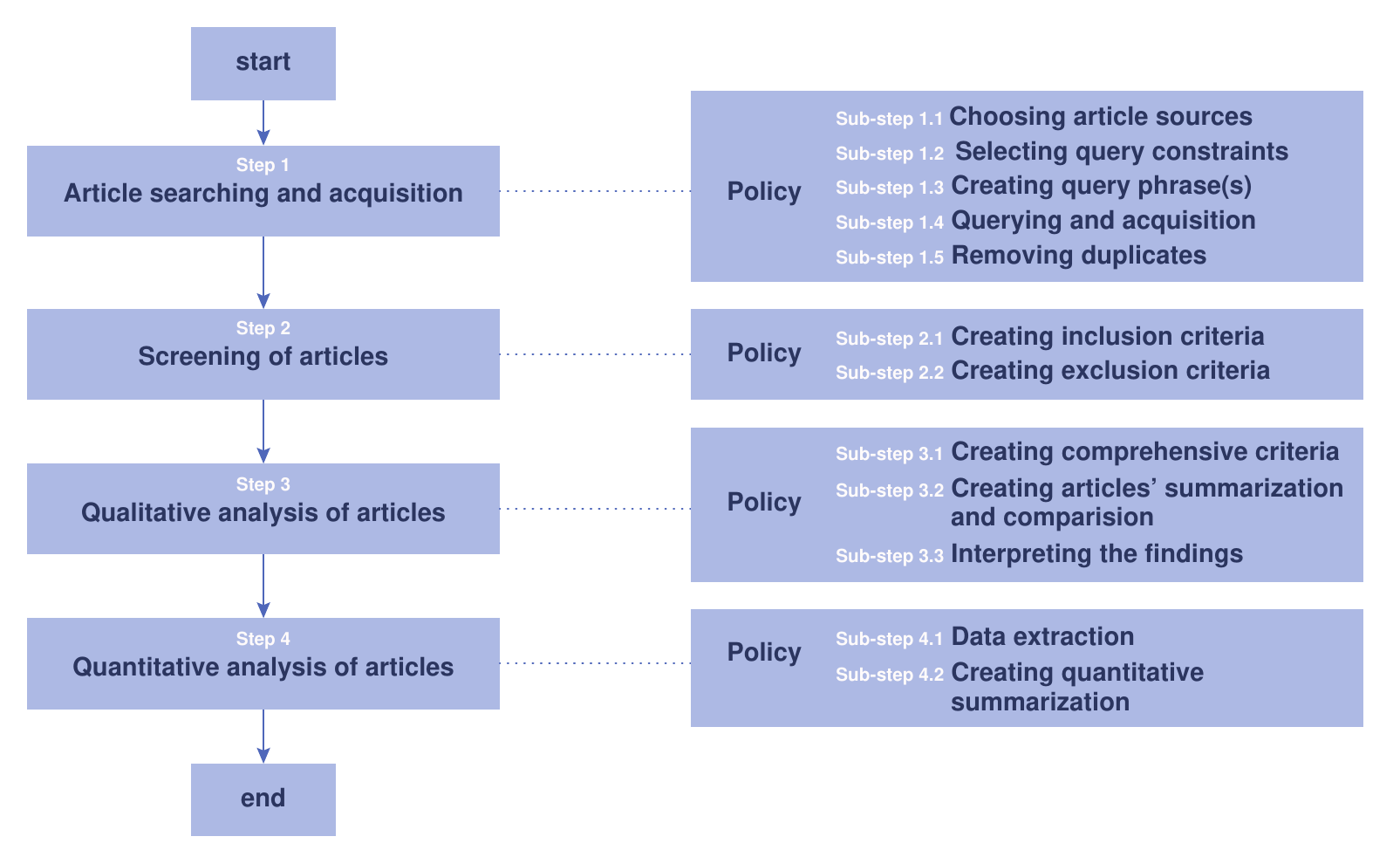}
\caption{A graphical representation of the review method for the studies.}\label{fig:ReviewMethod}
\end{figure}

Our review method comprises four phases: (1) article searching and acquisition; (2) screening of articles; (3) qualitative analysis of articles; and (4) quantitative analysis of articles. Figure~\ref{fig:ReviewMethod} presents these phases and their sub-steps.

Figure~\ref{fig:ReviewMethod} illustrates the four phases of the review process. The left rectangles represent the key phases, while the right rectangles denote the sub-steps within each phase. The arrows indicate the flow of the process. The article searching and acquisition phase involves the selection of sources and query constraints, the definition of query phrases, the execution of the querying and acquisition process, and the removal of duplicate articles. The screening phase includes the establishment of inclusion and exclusion criteria. The qualitative analysis phase involves the establishment of comprehensive criteria, the summarisation and comparison of articles, and the interpretation of the findings. Finally, the quantitative analysis phase comprises the extraction of data from each selected article to create statistics and quantitative summarisation to discover patterns. 

In our systematic review of target articles (those on information-fusion-based document classification), we used one abstract and citation database, Scopus, a well-known computer database that covers a wide range of (mostly peer-reviewed) research articles.  To target the most suitable articles, we formulated search criteria that enabled us to filter the results. We registered the number of records. After applying the screening policy, which enabled us to eliminate irrelevant articles in the context of our research question, we noted the final number of articles for further analysis.

Later, we developed a taxonomy that categorises the target articles systematically, facilitating structured analysis and synthesis. The taxonomy serves three key functions: (1) pattern and trend identification: by grouping studies based on document classification approaches, the taxonomy enables the detection of recurring themes and trends across the literature; (2) structured comparison: the taxonomy facilitates the synthesis of findings by providing a systematic framework for comparison; and (3) gap analysis: by organising studies into predefined categories, the taxonomy highlights under-represented areas of the research landscape, identifying gaps in the application of information fusion techniques for document classification.

The development of our taxonomy was an iterative, data-driven process based on principles of thematic analysis. We began by performing open coding on a representative subset of the selected articles, identifying key concepts related to modalities, fusion techniques, evaluation metrics, and more (see ~\ref{app:supplement} for the persistent link to the Supplementary Materials; within the Supplementary Materials, see Section \emph{Taxonomy of multimodal and multiview works} contains the full taxonomy). These initial codes were then  grouped iteratively into higher-level axial codes, which were refined continually against the full corpus of articles. This bottom-up, inductive approach ensures that the final taxonomic structure emerges directly from the evidence in the literature, rather than being imposed by pre-existing assumptions, thereby substantiating its novelty and data-driven nature.

\subsection{Review method: literature search and selection process}\label{sec:ReviewMethodLiteratureSearch}
Table~\ref{tab:queries} presents the steps of our article acquisition and exploration along with the final results of the number of articles selected for further analysis. Our search query combined keywords for information fusion (e.g. `multimodal', `multiview', and `information fusion') with terms for document classification (e.g. `text classification' and `document categorisation'). \ref{app:supplement} presents the exact query.

\begin{table}[H]
  \caption{Search strategy and results for article acquisition, 2025-01-03.}
  \centering
   \resizebox{0.7\linewidth}{!}{ 
    \begin{tabular}{p{6.5em}cp{19.5em}c}
\hline
 \thead{Search \\ source} & \thead{Number \\ of \\ records} & \thead{Screening \\ of \\ articles policy} & \thead{Number of articles \\ selected for \\ discussion and citation}\\
\hline    
\multicolumn{1}{c}{Scopus} &  374   & Inclusion criteria: study published in English, study focuses on document classification, study explicitly uses information fusion techniques, article is peer reviewed. Exclusion criteria: Study published in a language other than English, study not related to document classification, study does not use information fusion.& 139 \\
\hline
\end{tabular}%
}
\label{tab:queries}%
\end{table}%

\begin{figure}[H]
 \centering
  \includegraphics[scale=0.4,keepaspectratio]{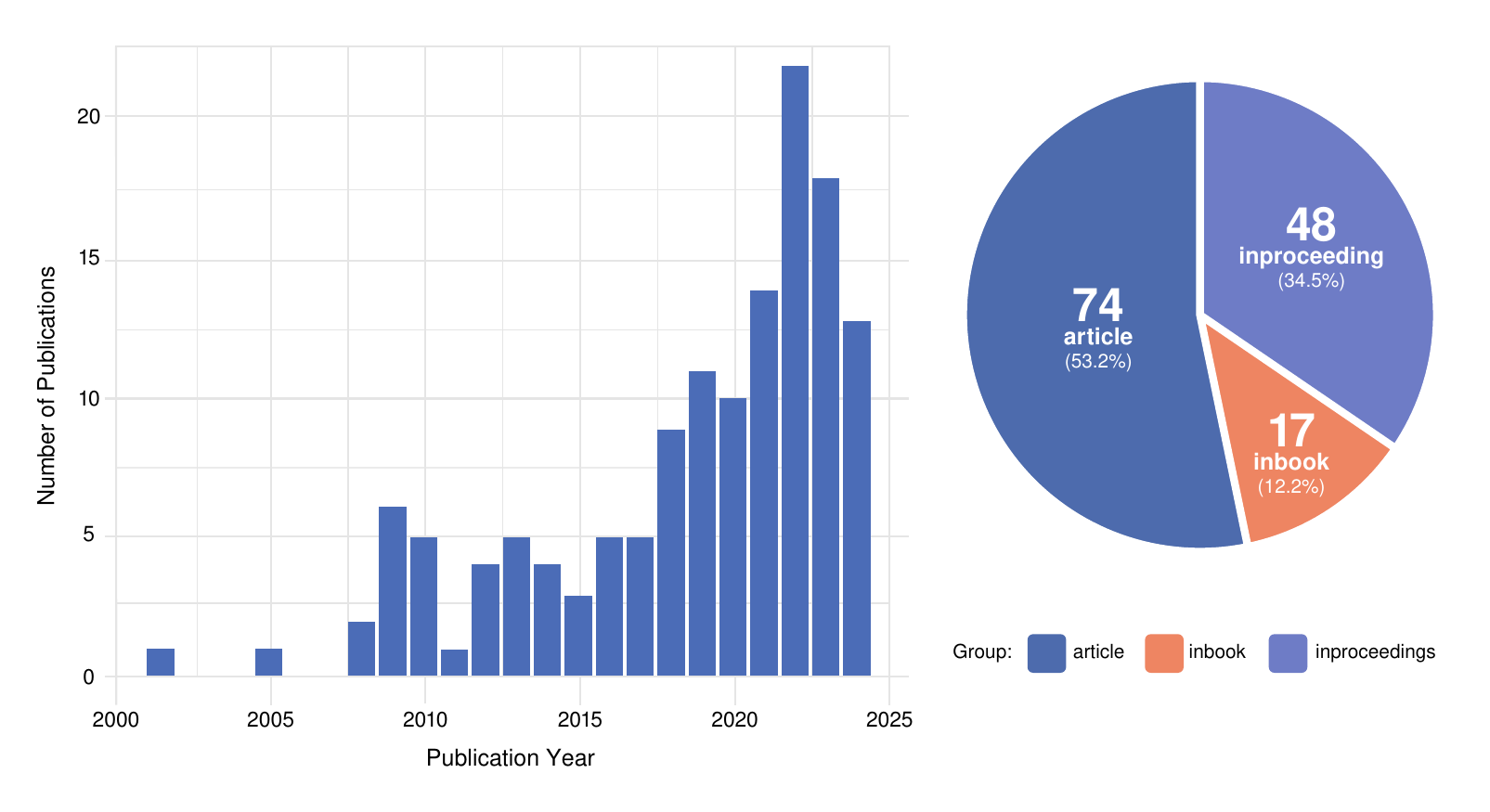}
 \caption{The left plot presents the number of publications per year from 2001 to 2024; the right plot presents the distribution of articles across different publication sources, including journals, books, and conference proceedings.}
\label{fig:pubdist}
\end{figure}

A total of 139 research articles were selected for the final analysis. Figure~\ref{fig:pubdist} presents the publication trends over the years and the distribution of the articles across various sources, including journals, books, and conference proceedings.

Analysing Figure~\ref{fig:pubdist}, we observe a consistent increase in the number of publications over the years, peaking in 2022, followed by a slight upward trend. Approximately half of the selected articles were published in journals, with the remainder distributed between conference proceedings and book chapters. Notably, the \textit{Lecture Notes in Computer Science} series (including sub-series such as \textit{Lecture Notes in Artificial Intelligence} and \textit{Lecture Notes in Bioinformatics}) was the most common publication venue, accounting for 12 articles. The journals include \textit{Neurocomputing} (five articles), \textit{Pattern Recognition} (five articles), \textit{Applied Sciences} (four articles), \textit{Expert Systems with Applications} (four articles), \textit{IEEE Access} (three articles), and \textit{Information} (three articles)~\footnote{Additional statistical details are available in the technical report accessible via Zenodo:
\url{https://doi.org/10.5281/zenodo.17141560}}.

\begin{figure}[H]
 \centering
  \includegraphics[scale=0.5,keepaspectratio]{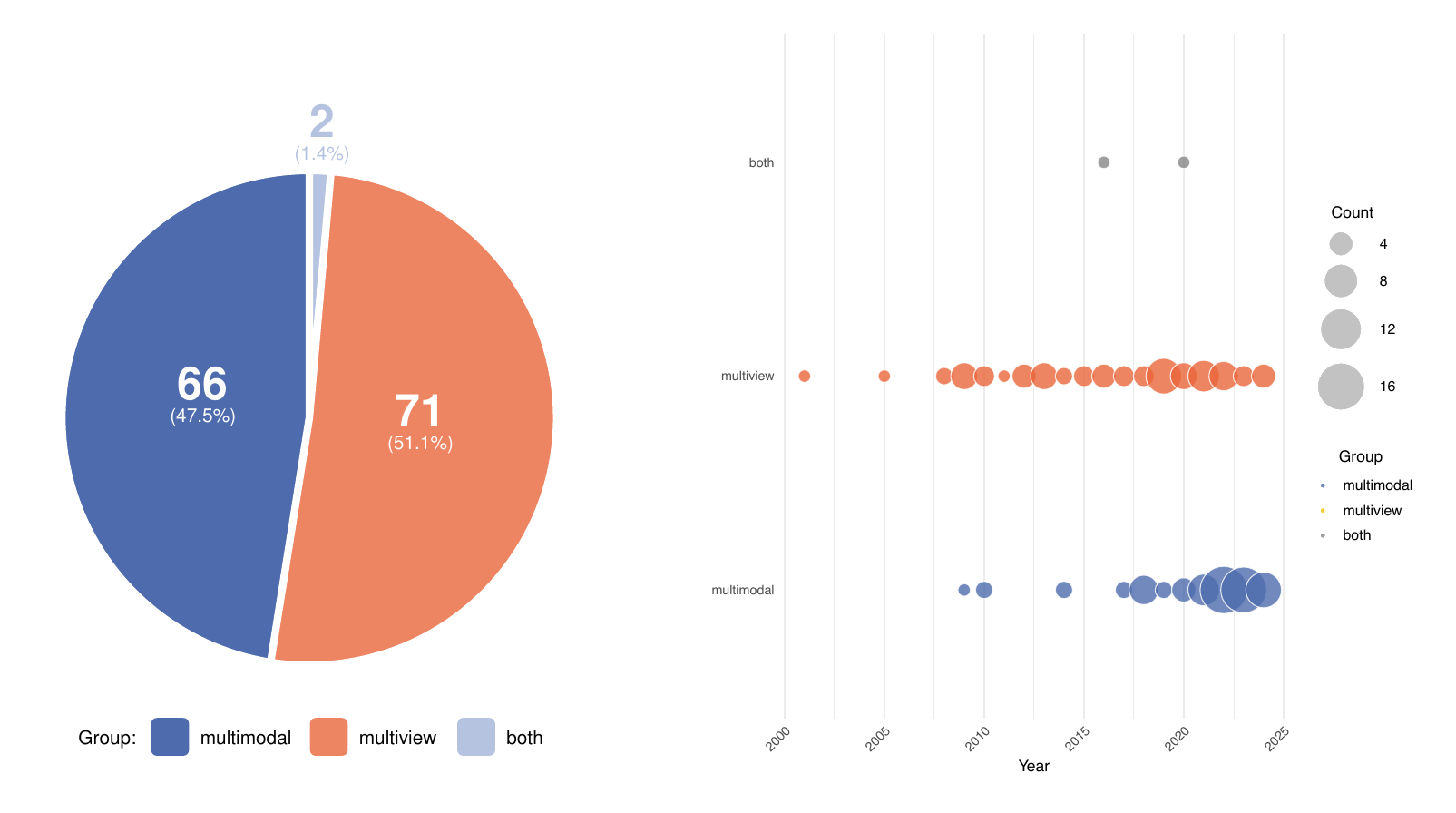}
 \caption{The left plot presents the number of publications on multimodal, multiview, and both learning approach; the right plot presents the time distribution of articles across those categories.}
\label{fig:piebuublemmmv}
\end{figure}

Ultimately, we selected 139 articles for further analysis, including 66 (47.5\%) on multimodal approaches, 71 (51.1\%) from the multiview learning scope, and two (1.4\%) that address both approaches. Figure~\ref{fig:piebuublemmmv} presents our collected articles' distribution and temporal trends. The pie chart on the left demonstrates a near equal distribution between multimodal and multiview articles, with only a minimal representation of articles that cover both approaches. The bubble plot on the right illustrates the temporal evolution of research in these areas. Notably, multiview learning publications began earlier (2001) and maintained a consistent presence throughout the study period; multimodal approaches emerged later (first appearing in 2009), but have increased in prominence since 2018. This temporal pattern suggests a growing interest in multimodal approaches in recent years, which potentially represents a shift in research focus in the document classification domain.

\section{Theoretical foundations}\label{sec:thfund}
Pattern recognition in document classification emerges from a hierarchical process that transforms raw data into actionable insights. The process begins with unprocessed documents in their original form, such as text files, PDFs, or web pages. A crucial preprocessing step, \textit{tokenisation}, is often performed. This process transforms raw text into a sequence of discrete units, or \textit{tokens}, such as words, sub-words, or characters. Such tokens serve as the foundational elements for subsequent processing.  

From the tokens, \textit{features} can be extracted: quantifiable properties that characterise particular aspects of the documents. \textit{Feature extraction}, involves identifying and measuring attributes that can range from simple counts like token frequencies to more complex properties like semantic relationships or indicators of writing style. Feature extraction can be based on various methods, including rule-based approaches and---more recently---data-driven techniques that involve \textit{feature learning}.  

In classical document classification methods, such as those based on the bag-of-words model, tokens obtained through tokenisation (typically words) were directly used as features. Each word was treated as a feature, and documents were represented as vectors that indicate the presence or frequency of those words. This direct correspondence meant that, in such methods, tokens and features were effectively synonymous.

After features are extracted, \textit{feature selection} or \textit{feature engineering} may be applied. Feature selection involves selecting the most relevant features for a given task, enhancing a model's efficiency and effectiveness. Feature engineering focuses on the creation of new features from existing ones, often leveraging domain knowledge to improve a model's performance.  

\textit{Feature learning}, in contrast to manual feature engineering, describes the automated learning of relevant features directly from raw or tokenised data using machine learning techniques. Neural networks, particularly transformer-based models, exemplify this approach, automatically constructing hierarchical, high-level features in their various layers. The tokenisation step provides a crucial initial representation upon which the network constructs higher-level features; this is beneficial for complex data with which manual feature engineering is challenging. The complexity of document classification necessitates a nuanced understanding of \textit{feature types}. Features exist on a spectrum from directly observable properties to derived characteristics: (1) \textit{Low-level features} comprise directly observable properties that can be extracted through straightforward parsing or pattern matching. They include document sections, paragraphs, character sequences, word occurrences, document length, and structural elements. Their extraction typically requires minimal computational inference; and (2) \textit{High-level features} represent abstract properties that must be derived through analysis or inference. Examples include semantic meaning, topical content, writing style characteristics, document categories, sentiment, author expertise level, and target audience identification. These features emerge through sophisticated analysis of low-level features and often require complex computational techniques for extraction.

To make these features computationally accessible and suitable for pattern analysis, they are encoded into \textit{representations}: formal structures that bridge the gap between raw data and computational analysis. A representation is a structured encoding of features. For instance, while `writing style' is a feature we can understand conceptually, its computational representation might take the form of a vector that encodes stylometric measurements. This transformation from features to representations enables systematic pattern analysis, and the representation then enables pattern recognition. While numerical representations form the computational foundation of document classification, the theoretical framework encompasses a broader spectrum of representation types, each of which serves particular purposes in document classification tasks: (1) \textit{Conceptual and cognitive representations} model human understanding and interpretation of information. They focus on capturing meaning, relationships, and context. Examples include semantic networks and knowledge graphs. These are important for tasks that require the deeper meaning of documents to be understood, such as fine-grained topic classification, cross-lingual document retrieval, and plagiarism detection based on semantic similarity; (2) \textit{Logical and structural representations} focus on the organisation and structure of documents. They capture hierarchical relationships, syntactic structure, and logical flow. Examples include parse trees and document outlines. These are useful for tasks like genre classification (e.g. research article vs. news article), automatic summarisation, and the identification of rhetorical structures for argument mining; and (3) \textit{Mathematical representations} encode information using numerical or mathematical structures, which enables computational processing. Examples include vectors (term frequency-inverse document frequency (TF-IDF), word embeddings), metrics (adjacency metrics for citation networks), and tensors. TF-IDF vectors are commonly used in traditional document classification algorithms like naïve Bayes or support vector machines (SVMs), while word embeddings are crucial for deep learning models such as recurrent neural networks (RNNs) and transformers.

These representation types form a natural hierarchy: conceptual models provide high-level understanding, logical structures formalise relationships, and mathematical representations enable computational processing. The translation between these layers presents a key challenge in document classification, as all representations must ultimately map to computational forms while preserving essential information from higher abstraction levels. In practice, effective document classification often requires combinations of multiple representation types to capture different aspects of document content. This multirepresentation approach enables robust pattern recognition by the leveraging of complementary perspectives on document structure and meaning. 

\begin{example}
A document could be represented by a logical structure (its outline) and a mathematical representation (word embeddings for each section).
\end{example}

Our theoretical framework develops this progression through two interconnected stages: (1) \textit{Information representation principles} (Section~\ref{sec:repform}): we establish how features are encoded into computational representations, providing the essential building blocks for pattern analysis; and (2) \textit{Pattern emergence and formalisation} (Section~\ref{sec:pattrep}): we examine how patterns manifest from these representations, with consideration for three key aspects: pattern manifestation across data structures, alignment with learning objectives, and the dynamic processes of pattern discovery.

\begin{figure}[H]
\centering
\includegraphics[scale=0.5,keepaspectratio]{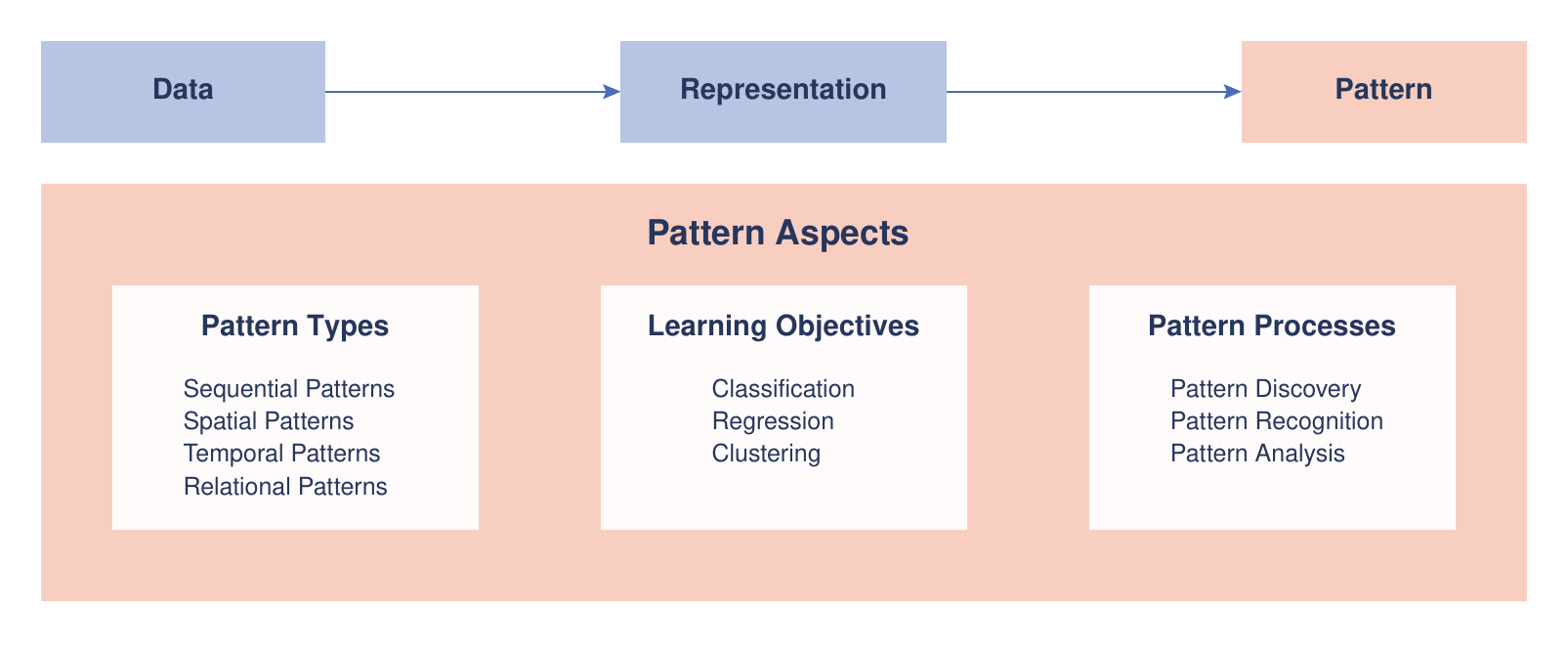}
\caption{The progression from raw data, through features and representations, to pattern recognition in document classification tasks. Features are extracted from raw data and encoded into representations, which then enable pattern processes.}
\label{fig:fundconc}
\end{figure}

Figure~\ref{fig:fundconc} presents this progression from raw data through features and representations to pattern recognition.

The progression from raw data to actionable insights, presented in Figure~\ref{fig:fundconc}, is fundamental for the advancement of document classification techniques. By comprehending the nature of documents and the most appropriate representations of their content, researchers can tailor patterns and create classification strategies to meet particular objectives, such as organising academic literature, managing digital libraries, or automating content categorisation.

\subsection{Representation formalisation}\label{sec:repform}
To establish a rigorous foundation for pattern recognition, we first formalise how features are encoded into representations. A representation, denoted as $R$, comprises three essential components $ R = (F, E, M)$, where: $F$ represents the format: the structural framework that organises the features (e.g. vector space for word frequencies, graph structure for citation networks); $E$ represents the encoding: a particular schema or method of transforming raw data into a structured, computationally useful format. This can include mathematical transformations or symbolic encodings (e.g. TF-IDF transformation, word embedding); and $M$ represents the meaning: the semantic interpretation that gives utility to the representation. Meaning connects the encoded data with the domain-specific understanding or task it serves (e.g. topic similarity, document relevance). For example, in document classification:

\[
R =
\begin{cases}
    F: & \text{Vector space } \textit{R}^n \text{ where } n = \text{vocabulary size} \\
    E: & \text{Term frequency-inverse document frequency (tf-idf)} \\
    M: & \text{Word importance within document context}
\end{cases}
\]

Representations, in turn, exist on a continuum from elementary to complex structures, rather than as discrete categories. This continuum reflects the transformation of representations based on analytical objectives. For instance, lexical frequency distributions—traditionally considered elementary representations—may reveal complex semantic patterns in some contexts, such as disciplinary vocabulary distributions in scholarly literature. Table~\ref{tab:exrepres} presents this spectrum, in which features serve as the raw or derived properties being measured, and representations provide the computational forms for encoding or measuring the features. 

\renewcommand{\arraystretch}{1.3}
\begin{tabularx}{\textwidth}{YYYYY}
\caption{A document representation framework that demonstrates how different representations are constructed from particular combinations of format (F), encoding (E), and meaning (M). The table illustrates the breadth of approaches in document classification.}
\label{tab:exrepres} \\

\toprule
\textbf{Representation} & \textbf{F (Format)} & \textbf{E (Encoding)} & \textbf{Relationship (E applied to F)} & \textbf{M (Meaning)} \\
\midrule
\endfirsthead

\caption[]{A document representation framework that demonstrates how different representations are constructed from particular combinations of format (F), encoding (E), and meaning (M). The table illustrates the breadth of approaches in document classification (continued).} \\
\toprule
\textbf{Representation} & \textbf{F (Format)} & \textbf{E (Encoding)} & \textbf{Relationship (E applied to F)} & \textbf{M (Meaning)} \\
\midrule
\endhead

\midrule
\multicolumn{5}{r}{\footnotesize\textit{Continued on next page}} \\
\endfoot

\bottomrule
\endlastfoot

Bag-of-words & Vector space \(R^n\) & Term frequency & Frequency count of terms in the vector space \(R^n\) & Word importance within a document \\
\midrule

TF-IDF vectors & Vector space \(R^n\) & Term frequency-inverse document frequency & TF-IDF weighting of terms in the vector space \(R^n\) & Word importance within a corpus \\

\midrule
Word embeddings & Vector space \(R^d\) & Neural network embedding (e.g. Word2Vec) & Dense vector representation of words in \(R^d\) & Semantic relationships between words \\

\midrule
Citation network & Graph & Adjacency matrix & Adjacency matrix representing connections in the graph & Relationships between documents based on citations \\

\midrule
Image pixel array & 2D Matrix & Raw pixel values & Matrix of raw pixel intensity values & Visual information in an image \\

\midrule
CNN feature vectors & Vector space \(R^k\) & Activations of a convolutional neural network & Vector of CNN activations in \(R^k\) & High-level visual features \\

\end{tabularx}
\renewcommand{\arraystretch}{1}

The synthesis of representations across this continuum enhances efficacy of classification. In scholarly article classification, character n-gram distributions might indicate disciplinary affiliations, while document structure analysis ranges from quantitative section analysis to rhetorical structure theory application. This hierarchical organisation of representation levels is exemplified in academic article analysis (Table~\ref{tab:fem}).

\begin{table}[H]
\centering
\caption{An example of hierarchical organisation of representation levels.}\vspace{-10pt}
\begin{tabularx}{\linewidth}{@{} l Y Y Y @{}}
\toprule
 & \textbf{F (Format)} & \textbf{E (Encoding)} & \textbf{M (Meaning)}\\
\midrule
$R_{\text{abstract}}$ &
Linear text block &
Dense summary encoding &
Research overview meaning\\
$R_{\text{introduction}}$ &
Multiparagraph structure &
Problem--context encoding &
Research motivation meaning\\
$R_{\text{methods}}$ &
Structured procedure format &
Technical detail encoding &
Methodological approach meaning\\
$R_{\text{document}}$ &
Hierarchical composition of section formats &
Integrated section encodings &
Complete research contribution meaning\\
\bottomrule
\end{tabularx}\label{tab:fem}
\end{table}

The formalisation through $R=(F,E,M)$ provides a unified framework for understanding how features at various levels contribute to the classification process, while acknowledging that the distinction between levels often depends on the analytical context and classification objectives. The $R=(F,E,M)$ framework itself formalises the process of representation construction, which describes the encoding of features into structured, computational formats. However, the particular encoding ($E$) within this framework can be determined in different ways. When the encoding is designed manually or through rule-based methods, it is called explicit representation construction; when the encoding is learned automatically from data using machine learning techniques, it is called representation learning. As we proceed to examine pattern emergence in Section~\ref{sec:pattrep} of this review, the dynamic nature of representation levels---whether constructed explicitly or learned automatically---provides the necessary foundation for understanding how patterns manifest across different analytical frameworks. The interplay between these levels influences the types and complexity of patterns we can discover directly.

\subsection{Pattern formalisation and types}\label{sec:pattrep}
The terms \textit{pattern recognition}, \textit{pattern identification}, and \textit{pattern discovery} are often used interchangeably to describe the process of identifying and characterising regularities in data. In the context of document classification, this process emerges from the interplay between features and their representations. Features are the measurable properties of documents, while representations are the computational structures that encode them. The selection of representation determines which patterns become discoverable in the classification process. For instance, representing text as a bag-of-words might reveal lexical frequency patterns but obscure sentence-level contextual patterns, whereas sentence embeddings could uncover relationships between ideas. Similarly, recognising citation networks requires graph-based representations that explicitly encode document-to-document relationships.

The theoretical framework of pattern representation in document classification rests upon three foundational concepts: features, representations, and patterns. Table~\ref{tab:fetrep} presents formal definitions of these key terms and their relationships in the document classification context.

\begingroup
  \small
    \begin{longtable}{@{} >{\RaggedRight}p{\PageFullWidthColOne} >{\RaggedRight}p{\PageFullWidthColTwo} >{\RaggedRight\arraybackslash}p{\PageFullWidthColThree} @{}}
    \caption{Core concepts of pattern recognition in document classification: the progression from features through representations to patterns. Each concept is illustrated with examples that are relevant to document classification tasks.}
    \label{tab:fetrep} \\
    \toprule
    \textbf{Term} & \textbf{Definition} & \textbf{Examples in document classification} \\
    \midrule
    \endfirsthead
    
    \caption[]{Core concepts of pattern recognition in document classification: the progression from features through representations to patterns. Each concept is illustrated with examples that are relevant to document classification tasks (continued).}\\ 
    \toprule
    \textbf{Term} & \textbf{Definition} & \textbf{Examples in document classification} \\
    \midrule
    \endhead
    
    \midrule
    \multicolumn{3}{r}{\footnotesize\textit{Continued on next page}} \\ 
    \endfoot
    
    \bottomrule
    \endlastfoot 

Feature 
    & A measurable  conceptual property or attribute of the document being measured (e.g. semantic meaning, writing style, topic) that can be used for classification.
    & Term importance, semantic similarity, document sentiment, readability, topical relevance, authorial style, and structural organisation (e.g. Introduction, Methods, Results, and Discussion (IMRAD) structure) \\
\hline
Representation 
    & A structured, computational encoding of features derived from raw data that enables pattern analysis and the derivation of higher-level features. In other words, the \textit{data structure} or \textit{mathematical object} used to encode one or more features.
    & \textit{Basic structures:}
      Scalar (document length); Vector (word embeddings); Matrix (term-document metrics); Tensor (contextual embeddings)
    
      \textit{Complex structures:}
      Neural networks (BERT embeddings or intermediate activations in the network at different layers); Graph structures (citation networks); Tree structures (section hierarchies); Probabilistic models (topic distributions) \\
\hline
Pattern 
    & A recurring, interpretable structure in representations that supports classification decisions. 
    & Standard section ordering in research articles, characteristic word distributions in scientific material, citation patterns in academic communities, consistent formatting in technical documents \\
\hline

\end{longtable}
\endgroup 

The framework presented in Table~\ref{tab:fetrep} establishes a hierarchical relationship between these concepts: features serve as measurable properties of documents, representations provide mathematical encodings of those features, and patterns emerge as interpretable structures that facilitate document classification tasks.

Once features are encoded into structured representations, the regularities, or `patterns', embedded within such representations may be identified. These recurring, interpretable structures guide document classification decisions. Such patterns emerge from how features are encoded in our selected representations and directly influence classification outcomes. To formalise this concept, we define a pattern $P$ as: $P = (S, C, T)$, comprising: $S$ (structure): the configuration of the basic elements that form the pattern; $C$ (constraints): rules that delineate valid configurations; and $T$ (transformations): permissible modifications that retain pattern integrity. For example, in research article classification:

\[
P_{research\_paper} =
\begin{cases}
    S: & \text{Abstract → Introduction → Methods → Results → Discussion} \\
    C: & \text{Introduction precedes Methods} \\
    T: & \text{Section summarisation} \\
\end{cases}
\]

Finally, a model $\mathcal{M}=(P,R,RR)$ may be defined as the mechanism that implements the recognition rules ($RR$s) to find instances of $P$ within a representation $R$. In other words, $RR$s are criteria for pattern identification. The representation $R$ is not directly part of the pattern $P$; rather, the $RR$s operate on the representation to determine whether it contains an instance of the pattern. For example, a recognition rule could be a function that verifies whether the sequence of section headers in $R$ matches the structure defined in $P$, or a condition that checks for significant keyword overlap between an Abstract and an Introduction section.
        
This formalisation aids in the systematic categorisation and exploration of patterns, such as sequential or spatial patterns, within documents. Patterns in document classification are multifaceted, comprising several critical components that collectively define their structure and functionality: (1) \textit{Examples of structure}: Document structure incorporates characters, symbols, strings, phrases, sentences, paragraphs, headings, and chapters, embodying the hierarchical and sequential nature of textual information; (2) \textit{Examples of constraints}: Document constraints encompass grammatical rules, spelling accuracy, and coherence, ensuring that the text adheres to language standards and logical flow; (3) \textit{Examples of transformations}: Document transformations include paraphrasing, summarisation, and translation, facilitating content adaptation without significant loss of original meaning or context; and (4) \textit{Examples of recognition rules}: Document recognition rules comprise keyword matching, topic modelling, and sentiment analysis---techniques that are pivotal for categorising and understanding the underlying sentiments or themes of textual data.

To illustrate these concepts in practical terms, consider the process of classifying a collection of documents based on thematic content. Here, $S$ might represent the thematic structure indicated by keyword distributions, $C$ could encapsulate thematic relevancy standards, $T$ might involve synonym substitutions to maintain thematic consistency, and $RR$ could involve machine learning algorithms designed to match documents to themes.

Clarifying how the framework applies to both traditional machine learning and to end-to-end learning methods like neural networks is key. In traditional machine learning, the representation (R) is typically constructed independently of the classifier (RR). 

\begin{example}
For example, TF-IDF vectors (R) are created first, and then a naïve Bayes classifier (RR) is trained on those vectors. In this case, R and RR are distinct components.
\end{example}

Contrastingly, in end-to-end learning methods like neural networks, the representation and the classification function are learned jointly. The entire neural network, from input layer to output layer, constitutes the \textit{RR}s. The intermediate activations in the network at different layers can be interpreted as the implicit representations (\textit{R}) learned by the network. These representations are not defined explicitly beforehand, but emerge as a result of the training process. Thus, while the \textit{R} and \textit{RR} components remain present in the framework, they are intertwined and learned simultaneously in end-to-end models.

\begin{example}
For example, we can consider a recurrent neural network (RNN) used for document classification. The input to the RNN is a sequence of word embeddings (which themselves can be considered a form of R at a lower level). The RNN processes this sequence, and the hidden states at each time step capture contextual information about the words already seen. These hidden states are the implicit representations (R) learned by the RNN. The final hidden state, or combination of hidden states, is then passed to a fully connected layer (or other classification layer) that produces the classification output. This fully connected layer, along with the entire RNN architecture and its learned weights, constitutes the RR. The network learns to adjust its weights so that the hidden states (R) capture relevant information for the classification task, with the final layer (RR) mapping these representations to class labels. Therefore, in the RNN, the representation and the classification logic are learned together as part of the overall RR.
\end{example}

In document classification, patterns manifest in several fundamental forms, each serving different analytical purposes. \textit{Sequential patterns} capture the ordering of elements, such as the characteristic progression from Abstract through Methods to Conclusions in research articles. \textit{Spatial patterns} reflect the physical arrangement of document elements, exemplified by the standardised layout of conference articles. \textit{Temporal patterns} consider timestamps---for example, in news articles---that indicate publication dates and update histories. \textit{Relational patterns} emerge from connections between documents or document elements, as seen in citation networks in academic literature. Finally, \textit{Association patterns} identify relationships between different entities in the document corpus.

The relevance and utility of particular patterns depend on the learning objective. \textit{Classification tasks} typically leverage patterns of feature co-occurrence—for instance, how technical terminology clusters in scientific disciplines. \textit{Regression tasks} might focus on patterns of feature correlation, such as how document structure relates to readability scores. \textit{Clustering tasks} often exploit patterns of document similarity, revealed through shared characteristics in their representations. Additionally, tasks that involve \textit{association rule learning}  identify patterns that reveal relationships between different entities in the document corpus, such as co-occurring terms or linked documents.

Pattern recognition in document classification proceeds through three interrelated processes. \textit{Pattern discovery} identifies previously unknown regularities in document collections, such as emerging citation patterns in growing research fields. \textit{Pattern recognition} applies known patterns to classify new documents, as when identifying document types based on their structure. \textit{Pattern analysis} evaluates the significance of identified patterns, helping to determine which patterns support classification decisions most effectively.

\begin{example}
For example, a scientific article classification system could combine sequential patterns (section order), relational patterns (citation networks), co-occurrence patterns (technical vocabulary), and spatial patterns (formatting) to achieve robust classification. Recognising the combined presence and strength of these patterns enables the accurate identification of research article type and research area.
\end{example}

\subsection{Summary}
When all of the above elements are combined, we observe that in our proposed framework, the \textit{recognition rules} (\( RR \)) component serves as the computational mechanism---the \textit{model}---that detects instances of patterns in the data representations \( R = (F, E, M) \). \( RR \) embodies the logic designed to identify particular \textit{structures} (\( S \)), while adhering to the defined \textit{constraints} (\( C \)) and accounting for permissible \textit{transformations} (\( T \)). \textit{Learning} is the process by which the \( RR \)s are trained or optimised to effectively identify instances of the defined patterns. This involves adjusting parameters within the \( RR \)s based on training data, and this can be performed using supervised, unsupervised, semi-supervised, or self-supervised methods.

\textit{Pattern recognition} is the application of the trained \( RR \)s to new data representations \( R \) to identify and interpret instances of the defined patterns. The \( RR \)s operate on the encoded features (\( E \)) within the representation \( R \), checking whether the data conforms to the defined structure (\( S \)), respects the constraints (\( C \)), and can be derived through permissible transformations (\( T \)). In summary, the \( P = (S, C, T) \), $\mathcal{M}=(P, R, RR)$ framework enables a comprehensive way of understanding how patterns are defined, learned, and recognised in document classification. The \( RR \) component corresponds to the model used for pattern recognition, and \textit{learning} is the process by which that component is trained. Crucially, the \( RR \) component does not exist in isolation; it is linked intrinsically to \( S \), \( C \), and \( T \), which collectively define the patterns it is designed to recognise. This integrated perspective underscores the interplay between patterns, models, and learning processes, highlighting how effective document classification relies on the harmonious alignment of these elements to identify and utilise patterns in data representations.

Moreover, the framework highlights the crucial interplay between representations and patterns. While the choice of representation influences which patterns can be discovered, the desired patterns (e.g. semantic relationships, citation networks) also inform the design of appropriate representations. This bidirectional relationship ensures that features, representations, and patterns align with the particular classification objectives. This alignment is essential for addressing complex document classification challenges, such as the organisation of digital libraries, the analysis of scholarly archives, and the automation of thematic categorisation. The framework naturally accommodates both multiview (multiple representations of the same data) and multimodal (representations of different data types) data, with the mechanisms for information fusion discussed in Section~\ref{sec:inffus}.

\section{Fundamentals of information fusion in pattern recognition}\label{sec:inffus}
Pattern recognition systems increasingly rely on diverse data representations to enhance their classification capabilities. Building upon the formal representation framework $R=(F,E,M)$ and pattern framework $P=(S,C,T)$ established in Section~\ref{sec:thfund}, this section explores how information fusion strategies leverage these foundations to combine multiple data sources effectively. Information fusion serves as a systematic approach to the integration of diverse representations, creating unified representations that produce more consistent, accurate, and comprehensive knowledge than individual sources alone can achieve. 

When combining different data sources, we must consider how their individual representations $R_1 =(F_1 ,E_1 ,M)$ and $R_2 =(F_2 ,E_2 ,M)$ can be integrated while preserving the shared meaning ($M$) relevant to the classification task. The fusion process creates a new representation $R_{fused} =(F_{fused} ,E_{fused} ,M)$, where $F_{fused}$ and $E_{fused} $ result from the combination of $F_1$ and $F_2$, and $E_1$ and $E_2$. 

Information fusion encompasses a broad spectrum of techniques and methodologies for the integration of diverse information sources. The scope of information fusion in pattern recognition can be categorised across multiple dimensions: (1) \textit{Source characteristics}: This considers the nature of information sources—whether they are homogeneous (the same type of data) or heterogeneous (different types of data)—and whether they offer complementary, redundant, or cooperative information. For instance, combining text with metadata enhances classification accuracy by providing complementary insights; (2) \textit{Data level integration}: This encompasses raw data combination, abstract representation fusion, and feature extraction methods. In document classification, the integration of word embeddings with structural document features may create richer representations for improved classification performance; (3) \textit{Model prediction level integration}: This involves combining predictions from different models, often through ensemble learning techniques. For example, aggregating outputs from naïve Bayes and SVM classifiers can enhance overall accuracy compared to individual models; and (4) \textit{Temporal dimension}: This addresses how information is integrated over time, including static (simultaneous integration), dynamic (sequential integration), and continuous fusion (real-time integration). Dynamic fusion has proved effective in processing streaming document collections in which data arrives sequentially.

In this framework, two fundamental learning paradigms have emerged: multimodal learning  and multiview learning. Both approaches aim to enhance pattern recognition through data combination, but differ significantly in their methodology and application.

\begin{figure}[H]
\centering
\includegraphics[scale=0.5,keepaspectratio]{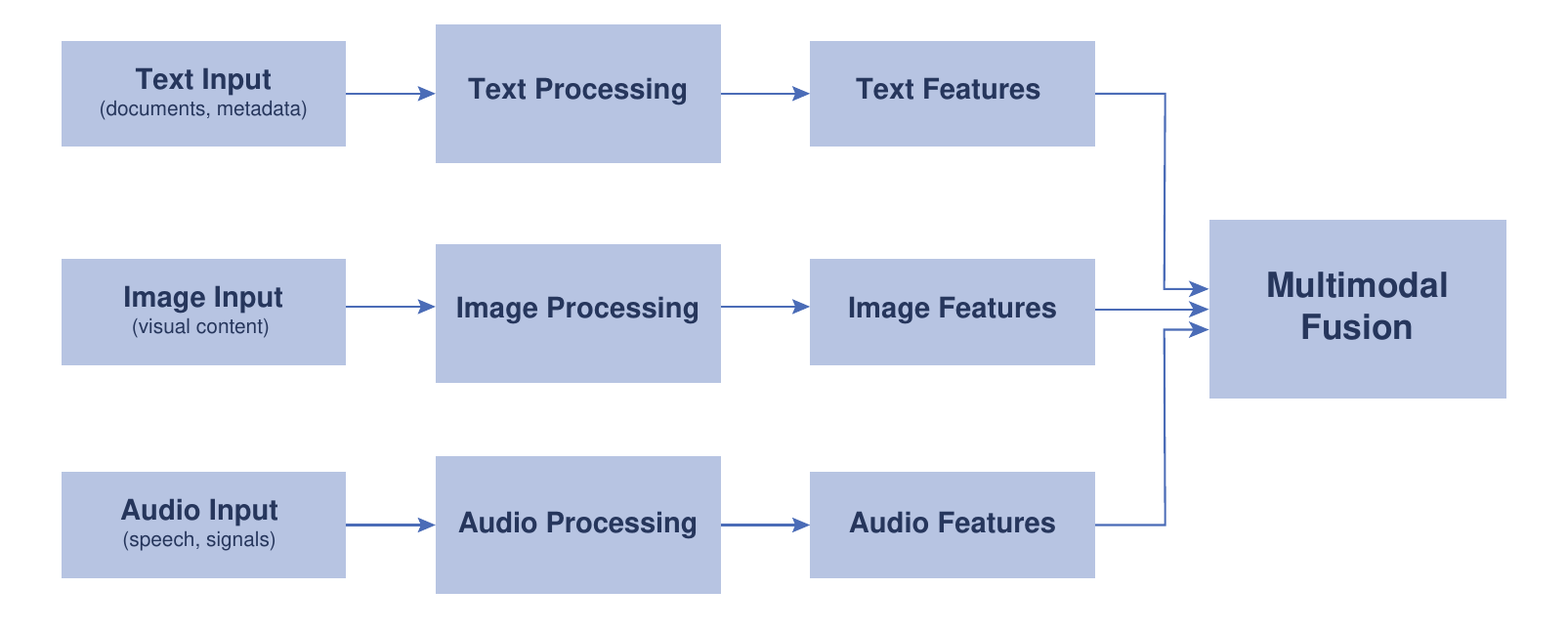}
 \caption{A multimodal learning framework that illustrates the integration of different data modalities (text, image, audio) through modality-specific processing paths before fusion.}
\label{fig:mml}
\end{figure}

\paragraph{Multimodal learning} combines data from fundamentally different modalities, such as text and images (Figure~\ref{fig:mml}). Each modality captures unique aspects of the information space. Multimodal fusion combines representations $R_1 =(F_1 ,E_1 ,M)$ and $R_2 =(F_2 ,E_2 ,M)$, where the meaning ($M$) relates to the same classification task, but the formats ($F1 \neq F2$ ) and/or encodings ($E1 \neq E2$ ) differ. 

Consider a document that contains both text and an image: 
\begin{itemize}
\item \textit{Text representation}: $R_{text} =(F_{vector} ,E_{TF-IDF} ,M_{topic\_relevance})$, where $F_{vector}$ is a  vector space format; $E_{TF-IDF}$ is an encoding using TF-IDF; and $M_{topic relevance}$ captures word importance in the document context relevant to the topic.    

\item \textit{Image representation}: $R_{image} =(F_{matrix} ,E_{CNN} ,M_{topic\_relevance})$, where $F_{matrix}$ is a matrix format representing pixel intensities; $E_{CNN}$ is an  encoding using a convolutional neural network (CNN); and $M_{topic relevance}$ captures visual features relevant to the topic.

\item \textit{Fusion process}: The fused representation is $R_{fused} =(F_{fused} ,E_{fused} ,M_{topic\_relevance })$, where $F_{fused}$ is a combined format; $E_{fused}$ is a combination of $E_{TF-IDF}$ and $E_{CNN}$, possibly through concatenation or another integration method.
\end{itemize}

By integrating these modalities, the fused representation captures both textual and visual information that is relevant to the classification task. 

\begin{figure}[H]
\centering
\includegraphics[scale=0.5,keepaspectratio]{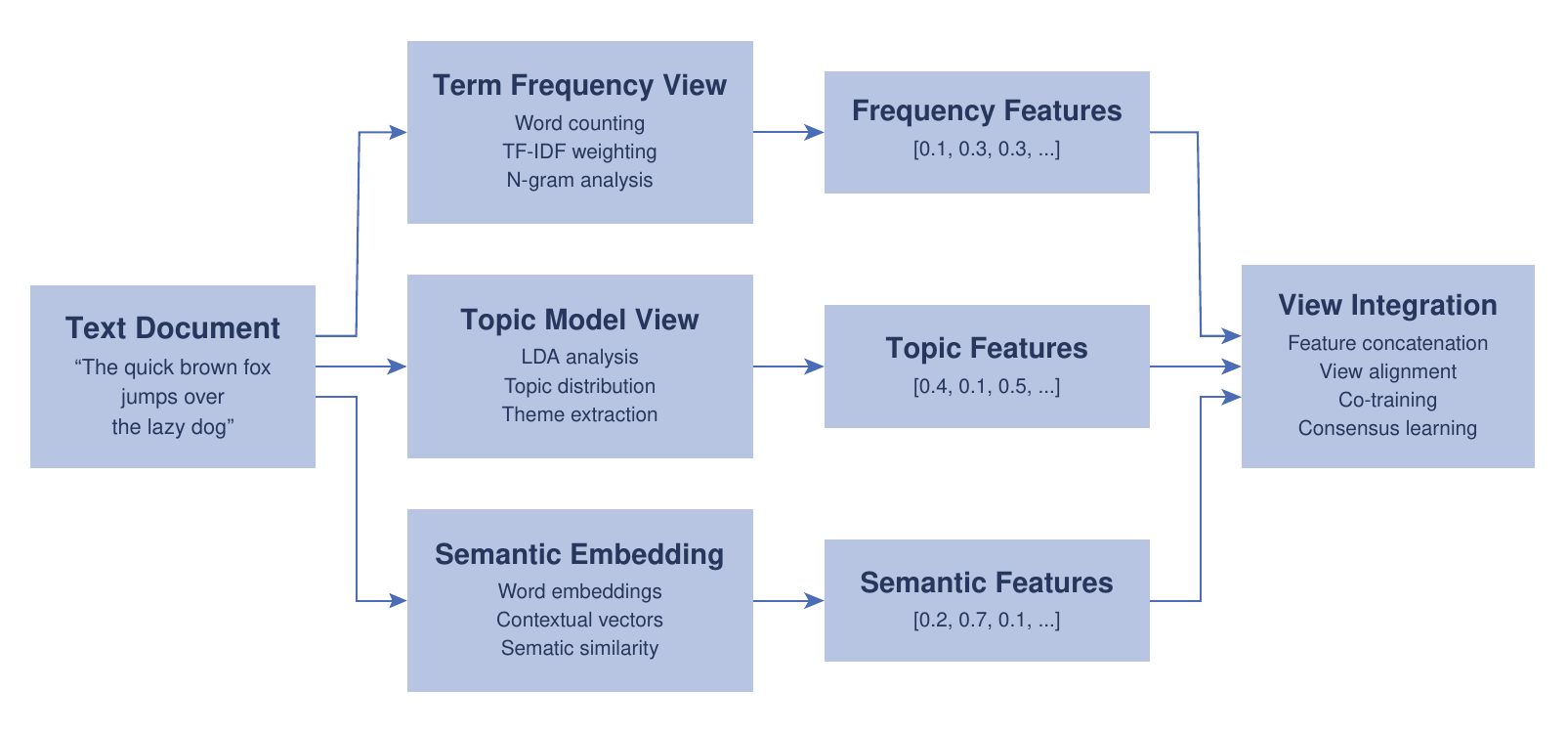}
\caption{A multiview learning framework that illustrates how different representations of the same data type are processed and combined for classification.}
\label{fig:mvl}
 \end{figure}

\paragraph{Multiview learning} operates on multiple representations of the same type of data, each offering a different perspective or feature set (Figure~\ref{fig:mvl}). The meaning ($M$) remains the same across views, while the formats ($F$) and/or encodings ($E$) vary. 

Consider a document represented in two different ways: 

\begin{itemize}
\item \textit{Bag-of-words representation}: $R_{BOW} =(F_{vector} ,E_{TF} ,M_{topic\_relevance})$, where $F_{vector}$ is a vector space format; $E_{TF}$ is an encoding using term frequency; and $M_{topic\_relevance}$ captures word importance based on frequency.
    
\item \textit{Word embedding representation}: $R_{WE} =(F_{vector}, E_{Word2Vec}, M_{topic\_relevance})$, where $F_{vector}$ is a vector space format; $E_{Word2Vec}$ is an encoding using a Word2Vec model; and $M_{topic\_relevance}$ captures semantic relationships between words.
    
\item \textit{Fusion process}: The fused representation is $R_{fused} =(F_{fused}, E_{fused} ,M_{topic\_relevance} )$, where: $F_{fused}$ is a vector space format; $E_{fused}$ is a  combination of $E_{TF}$ and $E_{Word2Vec}$, potentially through concatenation or dimensionality reduction techniques.
\end{itemize}

By combining these different views, the fused representation leverages both frequency-based and semantic information, enhancing robustness against noise and improving generalisation in classification tasks.  

\paragraph{Challenges in information fusion}

Several considerable challenges present opportunities for innovation in information fusion: (1) \textit{Data heterogeneity and incompatibility}: Different data sources may use incompatible formats or scales. The development of unified representations across heterogeneous data sources is crucial for effective fusion; (2) \textit{Temporal alignment and synchronisation}: In time-sensitive data, ensuring proper alignment between different sources is essential. Adaptive synchronisation techniques are necessary to handle temporal misalignments; and (3) \textit{Computational complexity and dimensionality}: The integration of multiple data sources can lead to high-dimensional feature spaces, increasing computational demands. Dimensionality reduction and efficient algorithms are needed to manage complexity while preserving important information. 

Advancements in representation learning and efficient fusion algorithms have begun to address these challenges, enabling more effective integration of diverse data sources. 
 
\subsection{Fusion strategies}
\begin{figure}[H]
 \centering
  \includegraphics[scale=0.5,keepaspectratio]{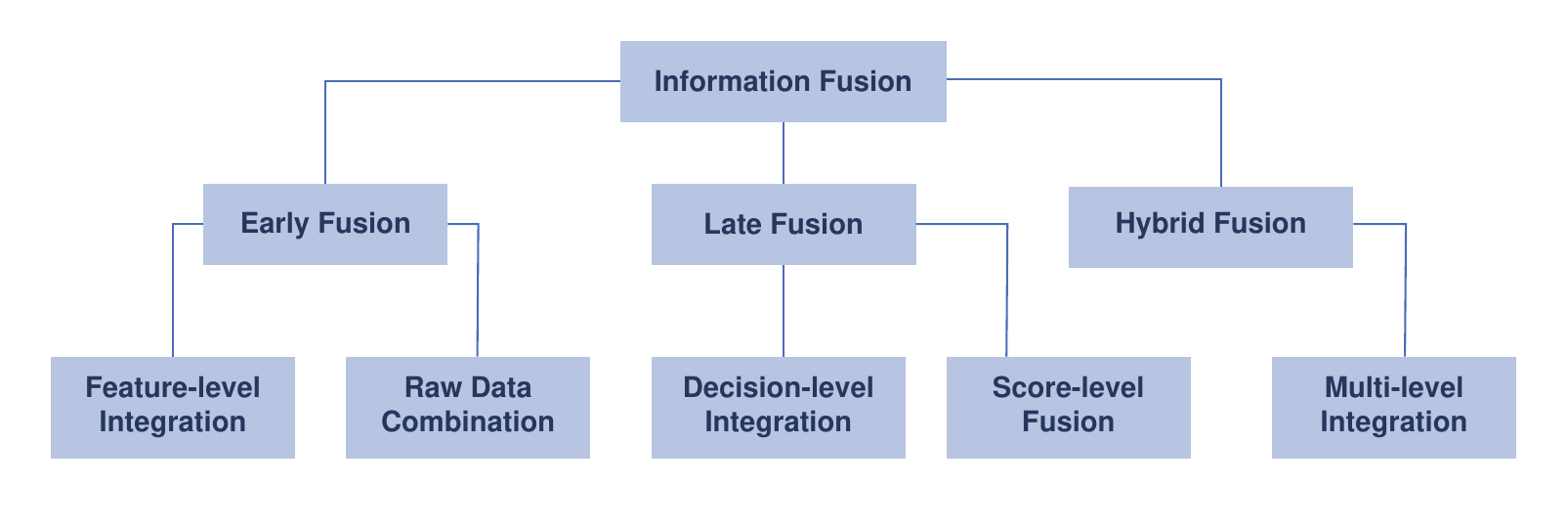}
 \caption{Fusion strategies connected to the $R=(F,E,M)$ framework: (a) early fusion combines encodings at the $E$ level; (b) late fusion combines classifier outputs at the $RR$ level; (c) hybrid fusion integrates features at both $E$ and $RR$ levels.}
\label{fig:fusstrat}
\end{figure}

Information fusion in pattern classification manifests through multiple strategic approaches, each operating at different levels of the recognition process. Such strategies can be connected explicitly to the $R=(F,E,M)$ framework. Figure~\ref{fig:fusstrat} presents these fusion strategies in the $R=(F,E,M)$ framework. 

\paragraph{Early fusion (feature-level fusion)}
This type of fusion occurs at the encoding level ($E$). 
The process combines encodings $E_1$ and $E_2$ to create $E_{fused}$, using concatenation, element-wise addition, or more complex integration methods. This results in the representation $R_{fused} = (F_{fused}, E_{fused}, M)$, 
where $F_{fused}$ depends on how $F_1$ and $F_2$ are combined, while $M$ remains consistent with $M_1$ and $M_2$.

\begin{example}
In document classification, text features (e.g. TF-IDF vectors) and metadata features (e.g. publication date encoded numerically) are combined into a single feature vector before classification.
\end{example}

\begin{example}
In neural networks, \textit{early fusion} describes the concatenation of input embeddings before the first processing layer or intermediate layers, i.e. the network merges multiple input encodings at the onset (e.g. concatenating feature maps) to produce a single unified representation for downstream classification.
\end{example}       

\paragraph{Late fusion (decision-level fusion)} 
This type of fusion occurs at the $RR$ level. The representations $R_1$ and $R_2$ are processed independently. 
Then, classifiers with recognition rules $RR_1$ and $RR_2$ generate separate outputs, which are subsequently combined to form the final decision. The fused recognition rule $RR_{fused}$ is a function of $RR_1$ and $RR_2$, with methods including majority voting, weighted averaging, or more sophisticated decision-combination techniques.
             
\begin{example}
Combining the predictions of a naïve Bayes classifier and an SVM trained on different feature sets.
\end{example}  
        
\begin{example}
In neural networks, the \textit{late fusion} model maintains separate branches—each processing its modality independently to generate separate class predictions, which are then combined post-classification via averaging, voting, or another fusion rule.
\end{example}  
     
\paragraph{Hybrid fusion} 
This type of fusion combines aspects of both early and late fusion. Some features are fused early at the encoding level ($E$), while others are processed separately and fused later at the recognition rule level ($RR$). This results in the hybrid representation and decision rules 
$R_{hybrid}$ and $RR_{hybrid}$, which incorporate both fused encodings and fused decisions.
      
\begin{example}
In sentiment analysis, text features and emoticon counts are fused at the feature level, while predictions from text and user behaviour classifiers are combined at the decision level.
\end{example}
        
\begin{example}
In neural networks, \textit{hybrid fusion} architectures blend early and late strategies by enabling intermediate feature interactions (e.g. cross-modal attention, shared transformer layers) before reaching separate classification heads that are fused at the final decision step.
\end{example}

The selection of fusion strategy often depends on the nature of the data, the requirements of the classification task, and the computational resources available. Each approach presents distinct advantages and challenges in terms of information preservation, computational efficiency, and model interpretability. 

\subsection{Alignment with classical fusion theory}
While modern document classification often employs learned fusion functions (e.g., neural networks), the strategies summarized in our framework can be related to classical information-fusion paradigms:

\begin{itemize}
\item Bayesian fusion and opinion pooling: Many late-fusion strategies in document classification combine the outputs of separate classifiers trained on different views (e.g., text, layout, metadata). When these outputs are calibrated posterior probabilities, simple averaging corresponds to a (possibly weighted) linear opinion pool, while multiplicative combination (under view-independence assumptions) is analogous to a logarithmic opinion pool. In our framework, the recognition rules ($RR$) for each view ideally produce view-conditional posteriors $\Pr(\text{class}\mid R_{\text{view}})$, which are then combined via Bayesian fusion operators to minimize expected decision risk.

\item Dempster--Shafer theory and evidential reasoning: Textual and multimodal documents frequently contain partially conflicting evidence (e.g., ironic text with a positive accompanying image). Whereas Bayesian models require a fully specified probability distribution over classes, Dempster--Shafer theory represents evidence via basic belief assignments (mass functions) and explicitly allows for uncertainty and ignorance. Within our framework, mass functions can be associated either with view-specific recognition rules or with higher-level patterns (e.g., segments, sections), and Dempster’s combination rule provides
a formal mechanism for aggregating evidence across views while quantifying conflict.

\item Kalman filtering and sequential fusion: When documents are processed sequentially (e.g., streaming text, incremental reading), recurrent architectures such as RNNs or Transformers with causal masking can be interpreted as learning a non-linear state-space model over the document. The hidden state plays a role analogous to the belief state in a Kalman filter: it is updated recursively as new observations (tokens, sentences, or multimodal cues) arrive, thus implementing temporal fusion of information within and across
views.

\item Probabilistic graphical models (PGMs): Classical multisensor and multiview fusion problems have often been formulated as PGMs, where the joint distribution over views and labels factorizes according to a graph structure. In our terminology, such models specify structural constraints on the pattern $P$ and the recognition rules $RR$ (e.g., conditional independence of views given the class, or view-specific latent variables). Naive Bayes, mixture-of-experts, conditional random fields and related models used for document classification can therefore be seen as particular graphical instantiations of our general fusion framework.
\end{itemize}

\subsection{Summary}
This section examines information fusion through the lens of the formal representation framework $R=(F,E,M)$. Information fusion brings together diverse data sources to create richer, more robust representations, improving classification performance and resilience against noise. Two central paradigms—multimodal learning and multiview learning—highlight the versatility in fusing different modalities or multiple representations of the same modality. Techniques range from early, feature-level fusion to late, decision-level fusion, as well as hybrid approaches that combine elements of both. Early, late, and hybrid fusion strategies entail different trade-offs depending on the data characteristics, classification goals, and computational constraints. The flexibility of this formal approach, which defines representations as $R=(F,E,M)$, patterns as $P=(S,C,T)$, and models as $\mathcal{M}=(P, R, RR)$, offers a structured way of comparing and adapting these fusion strategies, ensuring both performance gains and interpretability across diverse pattern recognition applications. Despite its benefits, information fusion also presents challenges including the management of heterogeneous data, the synchronisation of temporal information, and the control of high-dimensional feature spaces. Ongoing advances in representation learning and algorithmic efficiency continue to address these issues, paving the way for more flexible and powerful fusion strategies in pattern recognition.

\section{A qualitative analysis of multimodal and multiview document classification}\label{sec:qualanalysis}

\subsection{Multimodal representation approaches}
This section discusses the primary research questions, novel findings, and key challenges identified in the literature. It synthesises the major contributions of the works and highlights prominent open problems for future research. For a more granular classification, a detailed taxonomy that organises the studies by key characteristics—such as modalities, fusion techniques, and evaluation metrics—is provided in~\ref{app:supplement} for the link to the Supplementary Materials; within the Supplementary Materials, see Section \textit{Taxonomy of multimodal and multiview works}.

\subsubsection{Main research questions, novelty, findings, challenges and open problems}

\paragraph{Research questions and novelty}
Our thematic analysis presents a field in dynamic evolution. While text--image pairings remain the most common, and text--audio combinations are explored with increasing frequency, the research impetus extends far beyond simple modality fusion. Key motivations now centre on the development of trustworthy and selectively aligned representations \citep{zouh2023}. This addresses critical efficiency and accessibility concerns for real-world deployment \citep{garg2021, chenl2022}, as well as a persistent and critical inquiry that questions the genuine added value of multimodality \citep{hessel2020}. This is complemented by impactful applications in diverse domains like crisis management and healthcare. Moreover, emerging trends such as integration with large language models (LLMs) \citep{zhangy2024}, effective learning with limited data \citep{guelorget2021}, cross-lingual transfer, nuanced sequential understanding, and robustness to noise underscore a trajectory towards more sophisticated, reliable, and broadly applicable multimodal solutions. These research efforts have yielded several key insights, which are summarised below.

\paragraph{Findings} 
The findings and insights from the studies are summarised in Table~\ref{tab:multimodalkf}. These points highlight both the established benefits and the ongoing challenges in multimodal document classification, as well as prominent research trajectories. While these findings demonstrate clear progress, they also highlight a consistent set of practical and theoretical hurdles.

\begingroup
  \small
    \begin{longtable}{@{} >{\RaggedRight}p{\PageFullWidthColOne} >{\RaggedRight}p{\PageFullWidthColTwo} >{\RaggedRight\arraybackslash}p{\PageFullWidthColThree} @{}}
    \caption{Key findings in multimodal document classification.}
    \label{tab:multimodalkf} \\
    \toprule
    \textbf{Finding} & \textbf{Description} & \textbf{Citations} \\
    \midrule
    \endfirsthead
    
    \caption[]{Key findings in multimodal document classification (continued).}\\ 
    \toprule
    \textbf{Challenge} & \textbf{Description} & \textbf{Citations} \\
    \midrule
    \endhead
    
    \midrule
    \multicolumn{3}{r}{\footnotesize\textit{Continued on next page}} \\ 
    \endfoot
    
    \bottomrule
    \endlastfoot 

        Performance boost with multimodality &
        Combining text with other modalities (images, audio, or structured data) consistently improves classification accuracy (e.g. 5--15\% gains), particularly in complex domains. &
        \cite{andriyanovn2022}, \cite{guptad2018}, \cite{setiawan2021}, \cite{dacosta2022}, \cite{jiangs2024}, \cite{tengfeil2024}, \cite{braz2020}, \cite{luzdearau2023}, \cite{arlqaraleshs2024} \\
        \addlinespace

        Text dominance in unimodal settings &
        When individual modalities are compared, text-based classifiers often outperform those that rely solely on other single modalities (e.g. image or audio), particularly for semantic tasks. &
        \cite{matricm2018}, \cite{liparas2014}, \cite{rasheeda2023}, \cite{ortizperez2023}, \cite{akhiamov2018} \\
        \addlinespace

        Evolution of fusion strategies &
        Methodologies for combining modalities have advanced from simple concatenation to sophisticated techniques like attention, late fusion, and novel pooling/interaction architectures. &
        \cite{ravikiranm2019}, \cite{arlqaraleshs2024}, \cite{yushili2024}, \cite{liut2024}, \cite{tengfeil2024} \\
        \addlinespace

        Impact of pretrained models &
        Powerful pretrained language models (PLMs) are highly effective; incremental gains from other modalities can diminish, but techniques like cross-modal contrastive learning \& LLM+vision integration show promise. &
        \cite{ma2021}, \cite{chenl2022}, \cite{zhangy2024}, \cite{bakkalis2023} \\
        \addlinespace

        Strengths of multimodal approaches &
        Multimodal systems can: capture complementary information, offer robustness to noise/missing data, and show potential for cross-domain/lingual generalisation. &
        \cite{carmona2020}, \cite{guod2023}, \cite{wajdm2024}, \cite{liub2024}, \cite{luzdearau2023}, \cite{bakkalis2023}, \cite{debreuij2020} (note: \cite{fujinumay2023} highlights limits of cross-lingual generalisation) \\
        \addlinespace

        Weaknesses and limitations &
        Higher computational complexity, difficulty achieving true synergistic cross-modal interaction, class imbalance issues, and sensitivity to data quality (e.g. OCR errors). &
        \cite{jiangs2024}, \cite{anget2018}, \cite{hessel2020}, \cite{linckere2023}, \cite{sapeao2022}, \cite{jainr2019} \\
        \addlinespace

        Significant trends \& future directions &
        Adoption of contrastive learning, hierarchical document processing, focus on handling data imperfections, privacy-preserving learning, and growth in domain-specific applications. &
        \cite{zouh2023}, \cite{chenz2023}, \cite{tengfeil2024}, \cite{liut2024}, \cite{liub2024}, \cite{luzdearau2023}, \cite{garg2021}, \cite{braz2020}, \cite{reil2023}, \cite{chenl2022} \\

\end{longtable}
\endgroup 

\paragraph{Challenges and open problems} 
Common challenges, limitations, and potential areas for future research are presented in Table~\ref{tab:multimodalchallegng}.

\begingroup
  \small
    \begin{longtable}{@{} >{\RaggedRight}p{\PageFullWidthColOne} >{\RaggedRight}p{\PageFullWidthColTwo} >{\RaggedRight\arraybackslash}p{\PageFullWidthColThree} @{}}
    \caption{Key challenges in multimodal document classification.}
    \label{tab:multimodalchallegng} \\
    \toprule
    \textbf{Challenge} & \textbf{Description} & \textbf{Citations} \\
    \midrule
    \endfirsthead
    
    \caption[]{Key challenges in multimodal document classification (continued).}\\ 
    \toprule
    \textbf{Challenge} & \textbf{Description} & \textbf{Citations} \\
    \midrule
    \endhead
    
    \midrule
    \multicolumn{3}{r}{\footnotesize\textit{Continued on next page}} \\ 
    \endfoot
    
    \bottomrule
    \endlastfoot 
    
        Multimodal integration \& fusion &
        The effective combination of heterogeneous data types (e.g. text, image, layout, or audio) into a unified representation that captures synergistic information. This includes the selection of appropriate fusion strategies (early, late, hybrid) and the design of robust feature fusion modules. &
        \cite{wang2021}, \cite{yushili2024}, \cite{garg2021}, \cite{ghorbanali2024}, \cite{zhangx2010}, \cite{dongpin2022}, \cite{zouh2023}, \cite{adwaithd2022}, \cite{bakkalis2023}, \cite{chenz2023} \\ 
        \addlinespace
    
        Cross-modal alignment \& correspondence &
        The establishment of meaningful correspondences and semantic relationships between different modalities, particularly when dealing with varying granularities (e.g. image to sentence vs. image to document), implicit relationships, or sparse multimodal signals. &
        \cite{liut2024}, \cite{tengfeil2024} \\
        \addlinespace
    
        Data scarcity \& dataset creation &
        The considerable difficulty, cost, and time associated with creating, curating, and annotating large-scale, high-quality multimodal datasets, particularly for specialised domains, low-resource languages, or fine-grained tasks. This often necessitates the creation of in-house datasets. &
        \cite{garg2021}, \cite{wang2021}, \cite{setiawan2021}, \cite{ravikiranm2019}, \cite{chens2009}, \cite{ronghaop2024}, \cite{arlqaraleshs2024}, \cite{ortizperez2023}, \cite{yuet2022}, \cite{paraskevopoulos2022}, \cite{reil2023} \\
        \addlinespace
    
        Data quality, noise, and bias &
        Dealing with inherent issues in data such as label noise/errors, OCR inaccuracies, imbalanced class distributions, missing data, informal/mixed language, irrelevant content, and biases embedded in datasets, all of which can hinder model training, performance, and generalisation. &
        \cite{luzdearau2023}, \cite{fujinumay2023} (OCR), \cite{jainr2019} (OCR), \cite{andriyanovn2022} (text extraction), \cite{kanchid2022} (OCR), \cite{dacosta2022} (quality), \cite{liub2024} (intrinsic noise), \cite{wajdm2024} (noisy data), \cite{linckere2023} (noisy/unbalanced), \cite{sapeao2022} (transcription errors), \cite{reil2023} (quality, imbalance), \cite{jarquin2023} (imbalance), \cite{cristanim2014} (noise), \cite{chatziagapia2022} (ASR errors) \\
        \addlinespace
    
        Low-resource languages \& cross-lingual transfer &
        The adaptation of development of models for languages with limited linguistic resources (corpora, pretrained models) and datasets. The achievement of effective cross-lingual transfer of knowledge from high-resource languages, particularly to syntactically distant ones. &
        \cite{setiawan2021}, \cite{debreuij2020}, \cite{fujinumay2023}, \cite{ronghaop2024}, \cite{arlqaraleshs2024}, \cite{reil2023} \\
        \addlinespace
    
        Computational demands \& scalability &
        High computational requirements (CPU/GPU, memory) and extensive training time for complex multimodal models. Challenges in the deployment of large models efficiently, particularly on resource-constrained devices or for real-time applications. &
        \cite{gallo2021}, \cite{shah2023}, \cite{garg2021} (model size), \cite{ghorbanali2024} ( a solution to high computational demand), \cite{guptad2018}, \cite{anget2018}, \cite{jiangs2024}, \cite{wangq2022}, \cite{adwaithd2022} \\
        \addlinespace
    
        Hierarchical structure \& long document processing &
        The effective capture and modelling of the inherent hierarchical structure (e.g. sections, paragraphs, and sentences) and long-range dependencies in lengthy documents, particularly when integrating multimodal information and managing computational complexity. &
        \cite{tengfeil2024}, \cite{liut2024} \\
        \addlinespace
    
        Lack of standardised benchmarks \& fair evaluation &
        The absence of comprehensive, fair, and standardised benchmarks, datasets, and evaluation protocols makes it difficult to compare different approaches, measure true cross-modal contributions, and track progress reliably in the field. &
        \cite{hessel2020}, \cite{liub2024}, \cite{chens2009} \\
        \addlinespace
    
        Subjectivity, ambiguity, and annotation reliability &
        Handling tasks with inherent subjectivity (e.g. emotion, intent, or nuanced classifications) and addressing challenges in obtaining consistent and reliable annotations, including managing inter-annotator disagreement and the high cost of expert annotation. &
        \cite{wang2021} (emotion), \cite{guelorget2021} (slant, annotation effort), \cite{jarquin2023} (annotation, imbalance), \cite{kozienkop2023} (subjectivity, disagreement), \cite{rasheeda2023} (ambiguous designs), \cite{paraskevopoulos2022} (subjectivity) \\
        \addlinespace
    
        Model interpretability \& explainability &
        The development of models whose decision-making processes are transparent and understandable. This is crucial for debugging, building trust, ensuring fairness, and identifying whether models are genuinely leveraging multimodal signals or relying on unimodal cues. &
        \cite{hessel2020}, \cite{ortizperez2023} \\
        \addlinespace
    
        Domain-specific challenges \& adaptability &
        The need to tailor models and feature representations for particular domains (e.g. legal, medical, financial, e-commerce, or cultural heritage) that have unique vocabularies, structures, and data characteristics. Addressing issues like catastrophic forgetting when adapting to new domains/document types. &
        \cite{zingaro2021}, \cite{chenl2022}, \cite{chos2023}, \cite{guod2023} (ATC grammar), \cite{reil2023} (domain-specific vocabulary) \\
    
    \end{longtable}
\endgroup    
\subsubsection{Summary}
Our systematic review of multimodal representation approaches for document classification reveals several key insights: (1) \textit{Performance benefits with nuance}: While multimodal integration typically outperforms unimodal approaches \citep{jiangs2024, arlqaraleshs2024}, advantages can diminish when using powerful pretrained language models \citep{ma2021}. This suggests context-dependent utility; (2) \textit{Evolution of fusion techniques}: The field has progressed from simple concatenation to sophisticated architectures that incorporate attention mechanisms, cross-modal contrastive learning, and hierarchical document processing \citep{bakkalis2023, tengfeil2024, zouh2023}, with late or early fusion strategies remaining effective in many applications \citep{reil2023, kenny2023}; (3) \textit{Methodological trends}: We observe a clear shift from classical machine learning towards neural approaches, particularly with task-specific architectures that demonstrate stronger performance at the cost of reduced generalisation. Text--image combinations dominate (67.2\%), with increasing interest in audio and metadata integration~\footnote{see ~\ref{app:supplement} for the link to the Supplementary Materials; within the Supplementary Materials, see Section Multimodal representation approaches and Subsection \textit{Data modalities}}; (4) \textit{Research challenges}: Key challenges include the effective handling of data scarcity, computational demands, cross-modal alignment, and the hierarchical structure in long documents; and (5) \textit{Future directions}: Promising research trajectories include integration with LLMs \citep{zhangy2024}, contrastive representation learning \citep{bakkalis2023}, the handling of noisy multimodal data \citep{linckere2023}, and the development of benchmark datasets specifically designed to evaluate multimodal fusion strategies.

\subsection{Multiview representation approaches}
This section shifts its focus towards multiview representation, and likewise investigates the foundational concepts and state-of-the-art advancements in the field. This analysis aims to summarise the most impactful contributions and to outline unresolved issues that merit further investigation. As with the multimodal methods, these approaches are categorised in the detailed taxonomy provided in~\ref{app:supplement} for the link to the Supplementary Materials; within the Supplementary Materials, see Section \textit{Taxonomy of multimodal and multiview works}.

\subsubsection{Main research questions, novelty, findings, challenges and open problems}

\paragraph{Research questions and novelty} An analysis of the primary research questions posed in the surveyed articles reveals a field that is undergoing significant maturation and diversification. We observe a clear progression from foundational inquiries into basic feature combination (e.g. \cite{dasigiv2001}) towards explorations of sophisticated deep learning architectures, such as hybrid RNN models for time-series forecasting (e.g. \cite{gui2021}) and frameworks that feature attention mechanisms and CNNs for other tasks (e.g. \cite{liang2021, chens2019})—to achieve synergistic information fusion. These endeavours are often guided by the theoretical pursuit of maximising interview consensus while exploiting complementarity (e.g. \cite{jiax2021, pengj2018, hey2019}). Concurrently, the research questions reflect an increased focus on practical deployment challenges, evidenced by innovations in resource efficiency through semi-supervised, active, or transfer learning (e.g. \cite{jiz2024, maf2020, suns2010}), the pursuit of model robustness against adversarial attacks \cite{tianl2023} and distribution shifts \cite{lij2020}), and a growing emphasis on adaptive data representation (e.g. \cite{sus2021}). This evolution is mirrored in an expanding application landscape in which research questions tackle not only specialised difficult domains like short text (e.g. \cite{luox2022, karisanip2022}), cross-lingual transfer (e.g. \cite{guyo2012}), and extreme multilabel classification (e.g. \cite{chens2019}), but also a diverse array of tasks such as sentiment analysis \citep{zhang2021} and domain-specific problems in finance \citep{zhao2023} and healthcare \citep{huz2017}. Collectively, these trends, as articulated in the research questions, indicate a shift towards the development of more principled, data-aware, and resilient frameworks designed for effective real-world application. These research efforts have yielded several key insights, which are summarised below.

\paragraph{Findings} Table~\ref{tab:mvfindings} summarises the principal common themes, prevalent strengths, common weaknesses, and significant trends observed in the literature on multiview learning for document classification, based on the articles reviewed.

\captionsetup{type=table}
\begingroup
  \small
  \setlength{\tabcolsep}{4pt}
  \begin{longtable}{@{}>{\RaggedRight}p{0.35\textwidth}>{\RaggedRight\arraybackslash}p{0.65\textwidth}@{}}
    \caption{Key findings in multiview learning for document classification.}
    \label{tab:mvfindings} \\

    \toprule
    \textbf{Finding} & \textbf{Description} \\
    \midrule
    \endfirsthead

    \caption[]{Key findings in multiview learning for document classification (continued).} \\ 
    \toprule
    \textbf{Finding} & \textbf{Description} \\
    \midrule
    \endhead

    \midrule
    \multicolumn{2}{r}{\small\itshape continued on next page} \\
    \endfoot

    \bottomrule
    \endlastfoot

    \multicolumn{2}{@{}l}{\textbf{Common themes \& patterns}} \\
    \cmidrule(r){1-2}
    
    Performance superiority of multiview approaches &
    Across several studies, these methods consistently outperformed traditional singleview methods~\citep{zhang2021, bhatt2019, mmironczuk2020, pengj2018}. \\
    \addlinespace
    
    Integration of complementary information sources &
    Effectiveness stems from the leveraging of diverse, complementary information from different modalities or views~\citep{carmona2020, jiax2021}. \\
    \addlinespace
    
    Efficacy in cross-lingual \& cross-cultural applications &
    Multiview learning holds considerable promise for tasks that involve multiple languages, often by leveraging parallel corpora to learn shared representations~\citep{rajendran2016, aminim2010b, bhatt2019, guyo2012, aminim2010}. For nonparallel or comparable corpora, other methods use machine translation to generate the missing views, thereby creating a multiview setting~\citep{aminim2009}. \\
    \addlinespace
    
    Synergies with semi-supervised \& active learning &
    Multiview approaches combine effectively with semi-supervised or active learning paradigms to enhance classification with limited labelled data~\citep{karisanip2022, zhangx2009, jiz2024, maf2020, suns2010, suns2008, fakri2015, iglesias2016, gup2009}. \\ 
    \midrule

    \multicolumn{2}{@{}l}{\textbf{Prevalent strengths}} \\
    \cmidrule(r){1-2}
    
    Enhanced robustness to noise \& limited data &
    Multiview learning demonstrates improved resilience against noisy inputs, adversarial attacks, or scenarios with scarce labelled data or incomplete data~\citep{lij2020, jiz2024, zhangqi2024, doinychko2020, maf2020, fakri2015}.  \\
    \addlinespace
    
    Effective feature representation &
    These methods can generate more comprehensive and discriminative feature representations by combining information from different perspectives\citep{xuy2024, xuh2016, jiax2021, sus2021, chens2019, ferreira2018, longg2013}. \\
    \addlinespace
    
    Adaptability across domains \& tasks &
    Some approaches demonstrate versatility across various domains (e.g. social media, news, or scientific literature) or tasks (e.g. sentiment, topic, event detection, or troll detection)~\citep{samya2023, graffm2023, karisanip2022, tianl2023, xuc2017}. \\ 
    \midrule
    
    \multicolumn{2}{@{}l}{\textbf{Common weaknesses}} \\
    \cmidrule(r){1-2}
    
    Computational complexity \& training time &
    Multiview models often entail increased computational overhead and longer training durations due to their complex architectures~\citep{luox2022, liang2021}. \\ 
    \addlinespace
    
    Dependency on quality of individual views &
    The overall performance can be affected detrimentally by weaker or noisy views, particularly with insufficient labelled samples in one view~\citep{chenb2009}. \\
    \addlinespace
    
    Parameter sensitivity &
    Some approaches are sensitive to hyperparameter selection, requiring careful tuning for optimal performance~\citep{wangh2019, yangp2014}. \\
    \midrule
    
    \multicolumn{2}{@{}l}{\textbf{Significant trends}} \\
    \cmidrule(r){1-2}
    
    Integration of deep learning with multiview approaches &
    A clear trend towards combining deep learning architectures (e.g. GNNs or attention) with multiview frameworks for enhanced representation learning~\citep{xuy2024, zhangqi2024, zhang2021, huc2021, liang2021, lij2020, chens2019, xuh2016, jiz2024, varmanp2023, fengz2024}. \\ 
    \addlinespace
    
    Shift towards learned rather than handcrafted views &
    Recent research favours the learning of view representations directly from data over reliance on manually designed or heuristic-based views~\citep{samya2023, doinychko2020}. \\
    \addlinespace
    
    Addressing real-world limitations &
    Newer methods increasingly tackle practical challenges such as missing views, adversarial robustness, data imbalance, or knowledge loss over time (continual learning) ~\citep{doinychko2020, lij2020, liuj2022, zhangqi2024, tianl2023}. \\ 

\end{longtable}
\endgroup

\paragraph{Challenges and open problems} Some of the common challenges, limitations, and potential areas for future research are highlighted in Table~\ref{tab:mvchalanges}. While these findings demonstrate clear progress, they also highlight a consistent set of practical and theoretical hurdles.

\captionsetup{type=table}
\begingroup
  \small
  \setlength{\tabcolsep}{4pt}
  \begin{longtable}{@{}>{\RaggedRight}p{0.35\textwidth}>{\RaggedRight\arraybackslash}p{0.65\textwidth}@{}}

    \caption{Key challenges in multiview document classification.}
    \label{tab:mvchalanges} \\

    \toprule
    \textbf{Challenge} & \textbf{Description} \\
    \midrule
    \endfirsthead

    \caption[]{Key challenges in multiview document classification (continued)} \\ 
    \toprule
    \textbf{Challenge} & \textbf{Description} \\
    \midrule
    \endhead

    \midrule
    \multicolumn{2}{r}{\small\itshape continued on next page} \\
    \endfoot

    \bottomrule
    \endlastfoot

    \multicolumn{2}{@{}l}{\textbf{Data-related challenges}} \\
    \cmidrule(r){1-2}
    
    Data scarcity \& annotation cost &
    Insufficient labelled data; annotation is labour-intensive, time-consuming, and requires domain expertise~\citep{jiax2021,lij2020,akhtiamov2019,liuj2022,tianl2023}. \\
    \addlinespace
    
    Data quality \& representation &
    Issues with sparse (particularly short text) data, noise, incompleteness or missing views; informal language on social media; quality of artificially generated views~\citep{luox2022,liuj2022,zhangqi2024,doinychko2020,liaox2015}. \\
    \addlinespace
    
    Class imbalance &
    Severely skewed class distributions—particularly in domain-specific applications—degrade model performance~\citep{huz2017,carmona2020,liuj2022}. \\
    \midrule
        
    \multicolumn{2}{@{}l}{\textbf{Model architecture \& complexity}} \\
    \cmidrule(r){1-2}
    
    Optimisation \& parameter tuning &
    Nonconvex objectives, many hyperparameters, and complex structures are challenging to optimise—particularly with limited data~\citep{zhangd2013,huc2021,zhao2023,bhatt2019,guyo2012}. \\
    \addlinespace
    
    Feature extraction \& cross-view representation &
    Limits of singleview features; integrating multiple granularities/modalities; handling redundancy vs. complementarity between shared and particular information~\citep{liang2021,xuc2017,jiax2021,chens2019}. \\
    \addlinespace
    
    Computational complexity, scalability \& efficiency &
    High compute and memory costs for complex models, high-dimensional data, and iterative training hinder practical deployment~\citep{wangh2019,pengj2018,luox2022,max2020}. \\ 
    \midrule
    
    \multicolumn{2}{@{}l}{\textbf{Multiview integration challenges}} \\
    \cmidrule(r){1-2}
    
    View integration, balancing \& dependency &
    Combining diverse views, balancing their contributions, avoiding saturation, ensuring cross-view consistency, and managing independence/dependency assumptions~\citep{hey2019,zhang2021, zhangqi2024, fakri2015, brefeldu2015}. \\ 
    \addlinespace
    
    Cross-domain \& cross-language transfer &
    Bridging domain gaps and view--consistency violations in cross-domain settings~\citep{yangp2014}, as well as coping with translation errors/loss, or view-‑consistency violations across domains~\citep{guyo2012,zhangb2013}.  \\
    \addlinespace
    
    Robustness to adversarial attacks \& noise &
    Vulnerability to adversarially crafted inputs (particularly text) and the impact of noisy/corrupted views~\citep{lij2020}, ~\citep{mmironczuk2020} (this work acknowledges that its authors' models are not resistant to poisoning attacks and that this is an area for future research). \\
    \midrule
    
    \multicolumn{2}{@{}l}{\textbf{Evaluation \& practicality}} \\
    \cmidrule(r){1-2}
    
    Lack of standardised benchmarks/metrics &
    Difficult to evaluate and compare models, particularly for novel multimodal or domain-adaptation tasks. Implicit in many, e.g. \citep{lij2020} noting no benchmark for toxic Chinese content \\ 
    \addlinespace
    
    Interpretability \& explainability &
    Understanding the decisions of complex fusion mechanisms with multiple information sources. Generally a deep learning challenge; implied by \citep{maf2020} on lacking an optimisation model to explain mechanisms \\
	\end{longtable}
\endgroup

\justifying 
\setlength{\parindent}{15pt} 

\subsubsection{Summary}
This comprehensive analysis of multiview representation approaches for document classification studies that span two decades reveals several critical insights: (1) \textit{Consistent performance advantages}: Multiview approaches demonstrate superior performance over singleview methods across diverse domains, with their effectiveness stemming from the leveraging of complementary information sources \citep{zhang2021, bhatt2019, mmironczuk2020}. The field demonstrates balanced development between general-purpose frameworks (58\%) and task-specific approaches (42\%), with multilingual text categorisation and web document classification the most extensively explored applications~\footnote{see ~\ref{app:supplement} for the link to the Supplementary Materials; within the Supplementary Materials, see Section Multiview representation approaches and Subsection \textit{Problem specificity}}; (2) \textit{Methodological evolution and fusion strategies}: Early fusion dominates (42\%), followed by late fusion (33.3\%) and sophisticated hybrid approaches (24.6\%)~\footnote{see ~\ref{app:supplement} for the link to the Supplementary Materials; within the Supplementary Materials, see Section Multiview representation approaches and Subsection \textit{Fusion strategies}}. We observe a clear progression from traditional machine learning techniques towards deep learning architectures, with emerging trends in attention mechanisms \citep{liang2021, zhang2021}, graph-based fusion \citep{fengz2024, jiz2024}, and neural network integration replacing classical probabilistic methods; (3) \textit{Learning paradigm diversity}: While supervised learning remains prevalent (47.3\%), the field increasingly embraces advanced paradigms including semi-supervised learning (25.7\%), active learning, transfer learning, and emerging approaches like contrastive learning \citep{samya2023} and meta-learning \citep{tianl2023}. This reflects adaptation to real-world deployment challenges~\footnote{see ~\ref{app:supplement} for the link to the Supplementary Materials; within the Supplementary Materials, see Section Multiview representation approaches and Subsection \textit{Learning paradigms}}; and (4) \textit{Research maturation and practical focus}: Research questions have evolved from foundational feature combination inquiries towards sophisticated explorations of deep learning architectures, resource efficiency, model robustness, and adaptive data representation. The field demonstrates particular strength in cross-lingual applications and scenarios with limited labelled data through semi-supervised and active learning synergies.

\subsection{Multimodal and multiview approaches summary}
The qualitative analysis demonstrates how the theoretical frameworks established in Sections~\ref{sec:thfund} and~\ref{sec:inffus} manifest in contemporary research practices. The progression from representations $R=(F,E,M)$ through pattern frameworks $P=(S,C,T)$ to information fusion strategies provides a lens for understanding both the achievements and limitations observed in current multimodal and multiview document classification approaches.

\subsubsection{Theoretical framework validation through empirical evidence}
The reviewed literature validates our theoretical propositions. The representation framework $R=(F,E,M)$ effectively captures the diversity observed across studies: format variations ($F$) range from vector embeddings transformed into 2D visual representations~\citep{gallo2021} to complex tensor structures in modern transformer architectures~\citep{tengfeil2024}; encoding strategies ($E$) span from basic TF-IDF to sophisticated cross-modal contrastive learning~\citep{bakkalis2023}; while meaning preservation ($M$) remains the critical constant across all fusion approaches, ensuring semantic consistency despite architectural complexity.

Similarly, the pattern framework $P=(S,C,T)$ manifests clearly in practical applications. In multimodal approaches, we observe how recognition rules ($RR$s) adapt to different structural constraints ($C$)—from rigid early fusion concatenation to flexible attention-based late fusion mechanisms. The transformation component ($T$) proves particularly relevant in cross-lingual multiview scenarios~\citep{bhatt2019, rajendran2016}, in which linguistic transformations must preserve semantic content across language boundaries.

\subsubsection{Information fusion strategy effectiveness and limitations}
This analysis reveals that the choice between early, late, and hybrid fusion strategies impacts both performance and computational complexity. This aligns with the theoretical predictions from Section~\ref{sec:inffus}. Early fusion dominates multimodal approaches (58\% of studies) due to its ability to create unified representations $R_{fused}=(F_{fused},E_{fused},M)$ that capture cross-modal interactions; however, this dominance comes with increased computational demands and reduced modularity.

Conversely, multiview approaches demonstrate more balanced distribution across fusion strategies, with early fusion (42\%), late fusion (33.3\%), and hybrid approaches (24.6\%) each serving distinct purposes. This distribution reflects the inherent characteristics of textual multiview scenarios, in which views often represent different linguistic or semantic perspectives of the same underlying meaning ($M$). This makes late fusion through independent recognition rules ($RR_1, RR_2, ..., RR_n$) more naturally applicable.

\subsubsection{Methodological evolution and theoretical implications}
The temporal analysis reveals a clear evolution from classical machine learning-based fusion towards sophisticated neural architectures, which reflects advances in $RR$s in the pattern framework. However, this progression also introduces new challenges: (1) \textit{Representation learning complexity}: Modern approaches increasingly blur the boundaries between explicit representation construction and learned encodings in $R=(F,E,M)$. Deep learning models simultaneously learn optimal encodings ($E$) and $RR$s, making it difficult to isolate the contribution of fusion strategies versus representation quality; (2) \textit{Cross-modal alignment}: The challenge of establishing semantic correspondence between different modalities relates directly to the ensurance of consistent meaning ($M$) across diverse formats ($F_1, F_2, ..., F_n$) and encodings ($E_1, E_2, ..., E_n$). Our analysis reveals that attention mechanisms and contrastive learning approaches~\citep{zouh2023, samya2023} provide promising solutions for the protection of semantic coherence; and (3) \textit{Scalability and efficiency}: The computational complexity challenges observed across 73.5\% of the multimodal studies and highlighted in multiview approaches reflect the fundamental tension between comprehensive representation ($R$) and efficient $RR$s.

\subsubsection{Bridging theory and practice: gaps identified}
This analysis reveals several gaps between theoretical potential and practical implementation: (1) \textit{Evaluation methodology}: The lack of standardised evaluation protocols (noted across both multimodal and multiview literature) suggests insufficient attention to meaning preservation ($M$) verification. Only 11.8\% of multimodal and 23.3\% of multiview studies employ statistical significance testing. This indicates inadequate validation of whether fusion truly improves pattern recognition or merely increases model complexity; (2) \textit{Challenges in reproducibility}: The low reproduction rates (26.5\% for multimodal, 16.4\% for multiview) suggest that many studies fail to provide sufficient detail on their representation construction process, particularly the encoding components ($E$) and ($RR$) implementation; and (3) \textit{Cross-domain generalisation}: While our theoretical framework predicts that properly constructed representations should generalise across domains when meaning ($M$) is preserved, the predominance of domain-specific datasets (61.1\% multimodal, domain-specific applications) suggests limited validation of this theoretical prediction.

\subsubsection{Emerging trends and advanced techniques}
Recent developments introduce innovative approaches that extend the theoretical framework: (1) \textit{Dynamic fusion architectures}: Attention mechanisms and adaptive weighting represent evolution in $RR$s that can adjust dynamically based on input characteristics, moving beyond static fusion strategies; (2) \textit{Large language model integration}: The emergence of LLM-enhanced approaches~\citep{zhangy2024} suggests new possibilities for encoding ($E$) in the representation framework, in which pretrained knowledge provides robust semantic foundations for meaning preservation ($M$); and (3) \textit{Contrastive and self-supervised learning}: These approaches address the alignment challenges in multimodal fusion directly by learning representations in which semantic similarity is preserved across modalities, operationalising meaning consistency ($M$) in the embedding space.

\subsubsection{Future directions and theoretical extensions}
Based on this synthesis of theoretical foundations and empirical evidence, several promising research directions emerge: (1) \textit{Principled fusion strategy selection}: The development of theoretical guidelines for choosing fusion strategies based on modality characteristics, data availability, and computational constraints, extending the $R=(F,E,M)$ framework with decision criteria; (2) \textit{Unified evaluation frameworks}: The creation of standardised benchmarks that explicitly test meaning preservation ($M$) across different fusion approaches, addressing the challenges in reproducibility identified in the literature; (3) \textit{Cross-modal semantic consistency}: The advancement of the theoretical understanding of how to maintain meaning ($M$) when combining fundamentally different formats ($F_1 \neq F_2 \neq F_3$), particularly in scenarios with more than two modalities; (4) \textit{Interpretable fusion mechanisms}: The development of approaches that make the transformation from individual representations to fused patterns transparent, enhancing the explainability of $RR$s in complex fusion architectures; and (5)\textit{Adaptive and context-aware fusion}: Exploration of how fusion strategies can adapt dynamically to data characteristics and task requirements, potentially through learned policies that optimise the entire $R=(F,E,M) \rightarrow P=(S,C,T) \rightarrow \mathcal{M}=(P, R, RR)$ pipeline.

\section{A quantitative analysis of multimodal and multiview document classification}\label{sec:quananalysis}

\subsection{A quantitative analysis of articles}\label{sec:QuantitativePolicy}
To conduct a comprehensive quantitative analysis that answers our main quantitative research question regarding the performance advantage of multiview/multimodal approaches (\textit{RQ3}), we extracted performance metrics from all eligible experiments identified in our literature search. This section quantitatively compares studies that use multiple sources of information or representation (i.e. multiview and multimodal) against those that use a single source (i.e. singleview and unimodal).

To ensure methodological rigour and comparability, we imposed strict eligibility criteria for the quantitative synthesis beyond those used for the broader systematic review. A study was included in the meta-analysis if and only if it: (1) reported results for at least one fusion-based model and a clearly defined non-fusion baseline evaluated on the same dataset and under the same evaluation protocol \emph{within that study}; (2) provided sufficient numerical detail to compute a study-level effect size, i.e., absolute performance scores (such as accuracy) for both fusion and baseline models rather than only relative improvements or graphical summaries; and (3) employed an evaluation protocol that enabled meaningful comparison, such as a held-out test set or cross-validation with identical splits. Different studies were therefore allowed to use different datasets, and their effect sizes were combined in the meta-analysis. Studies that discussed fusion conceptually or without comparable baselines were retained for the qualitative synthesis but excluded from the quantitative meta-analysis to prevent methodological artefacts.

Our investigation was a two-stage process. The first was an exploratory stage that established the presence and character of an effect. The second was a formal meta-analysis that derived the most precise estimate of that effect's magnitude\footnote{The full technical report, including the raw data and computational details, is available for review in our Zenodo repository:
\url{https://doi.org/10.5281/zenodo.17141560}}\label{foot:1}. This process was designed to answer four core research questions:

\begin{itemize}
\item \textit{RQ3.1:}  Is there a statistically significant difference in the mean reported outcomes across studies?
\item \textit{RQ3.2:} Is this finding robust to the influence of potential outlier studies?
\item \textit{RQ3.3:} What is the nonparametric magnitude of the difference between the groups that represents the probability of superiority?
\item \textit{RQ3.4:} What is the best estimate of the true average effect of the multiview/multimodal approach, after weighting each study by its statistical precision, and how consistent is that effect across the literature?
\end{itemize}

To address these questions, we performed all statistical analyses using the R Project for Statistical Computing \citep{rprojectBase}. 
We report heterogeneity statistics ($\tau^2$, $\tau$, $I^2$, and $H^2$) and 95\% prediction intervals in the Supplement (see ~\ref{app:supplement} for the link to quantitative analysis technical report). Where studies did not report variances, we derived standard deviations (SDs) as detailed in the technical report and ran analyses with we imputed the SD using the pooled average SD calculated from studies that did provide this information, a common practice in meta-analysis to enable their inclusion in a weighted model. We also assessed potential small-study effects (publication bias) using funnel plots and Egger’s regression test for outcomes with sufficient studies ($k \ge 10$). No significant funnel plot asymmetry was detected for Accuracy or F1-score in either multimodal or multiview analyses (see technical report for full outputs\footnotemark[7]). 

For all hypothesis tests, we used a significance level of $\alpha = 0.05$. We employed a permutation test on the set of calculated study-level mean differences to address \textit{RQ3.1}. The test directly evaluates whether the observed grand mean of these study-level differences is significantly different from zero---providing an initial answer as to whether, on average, a multiview/multimodal approach yields a different result. To address \textit{RQ3.2} and to assess the robustness of our findings, we applied a Wilcoxon signed-rank test~\citep{rprojectBase} to the same set of study-level mean differences. The purpose of this second test was to evaluate the central tendency (median), which is less sensitive to extreme values. A significant result here indicates that the effect is not driven by a few studies with unusually large means, but is instead a typical finding in the literature. The test thus serves as a critical robustness check. To answer \textit{RQ3.3}, we computed Cliff's delta ($\delta$)~\citep{rprojectEffSize}. This nonparametric effect size quantifies the size of the difference between the two groups in probabilistic terms. Specifically, it measures the difference by calculating the probability that a randomly selected outcome from a multiview/multimodal study is superior to one from a singleview/unimodal study, and vice versa. This provides an intuitive measure of effect size that is robust to outliers and makes no assumptions regarding the data's underlying distribution. To address \textit{RQ3.4}, we conducted our primary analysis using a robust random effects meta-analysis~\citep{rprojectMetafor}. The method synthesises the results from all studies to provide a single, overall estimate of an effect. Crucially, it employs inverse-variance weighting, which grants greater influence to studies that provide more precise estimates of the effect (i.e. those with lower variance). Such precision is typically achieved through larger sample sizes or lower within-study variability. This `gold-standard' approach yields the most reliable estimate of the true average effect size and its 95\% confidence interval, while also enabling us to assess the degree of consistency, or heterogeneity, across studies.

\subsection{Multimodal representation approaches}

\subsubsection{Evaluation details}

\begin{figure}[H]
 \centering
  \includegraphics[scale=0.4,keepaspectratio]{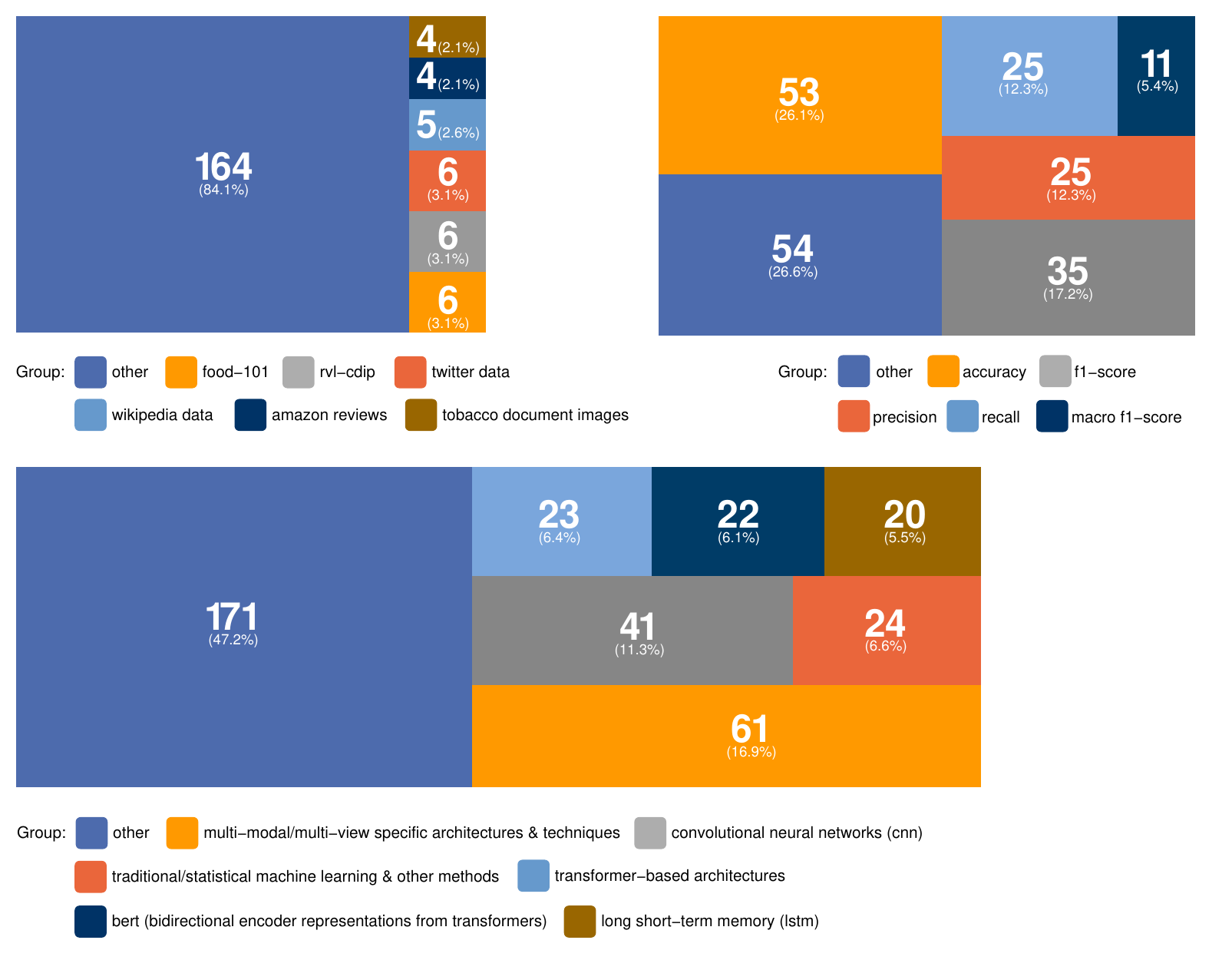}
 \caption{An overview of evaluation practices in multimodal experiments: The top-left plot presented the frequency distribution of the datasets; the top-right plot presents the frequency distribution of the evaluation metrics; and the bottom plot displays the distribution of the baseline models used for comparison.}
\label{fig:mmdsmetmod}
\end{figure}

\paragraph{Datasets, models selected for comparison, and performance metrics} 
The analysis of information-fusion-based document classification research reveals several patterns in the evaluation methodologies, which are presented in Figure~\ref{fig:mmdsmetmod}. Regarding datasets, researchers predominantly use diverse, specialised collections (84.1\% categorised as `other'), with standardised benchmarks like Food-101, RVL-CDIP, and Twitter data each appearing only in 3.1\% of occurrences. For model comparisons, while 47.2\% of baseline approaches fall into miscellaneous categories, multimodal/multiview-specific architectures (16.9\%) and CNNs (11.3\%) emerge as popular benchmarks, followed by traditional machine learning methods (6.6\%) and transformer-based models (6.4\%). Regarding evaluation metrics, while a diverse collection of `other' metrics (including computation time and memory usage) cumulatively represents the largest share (26.6\%), accuracy stands out as the most frequently used single performance metric (26.1\%), closely followed by F1-score (17.2\%), and precision and recall (both 12.3\%). 

\begin{figure}[H]
 \centering
  \includegraphics[scale=0.4,keepaspectratio]{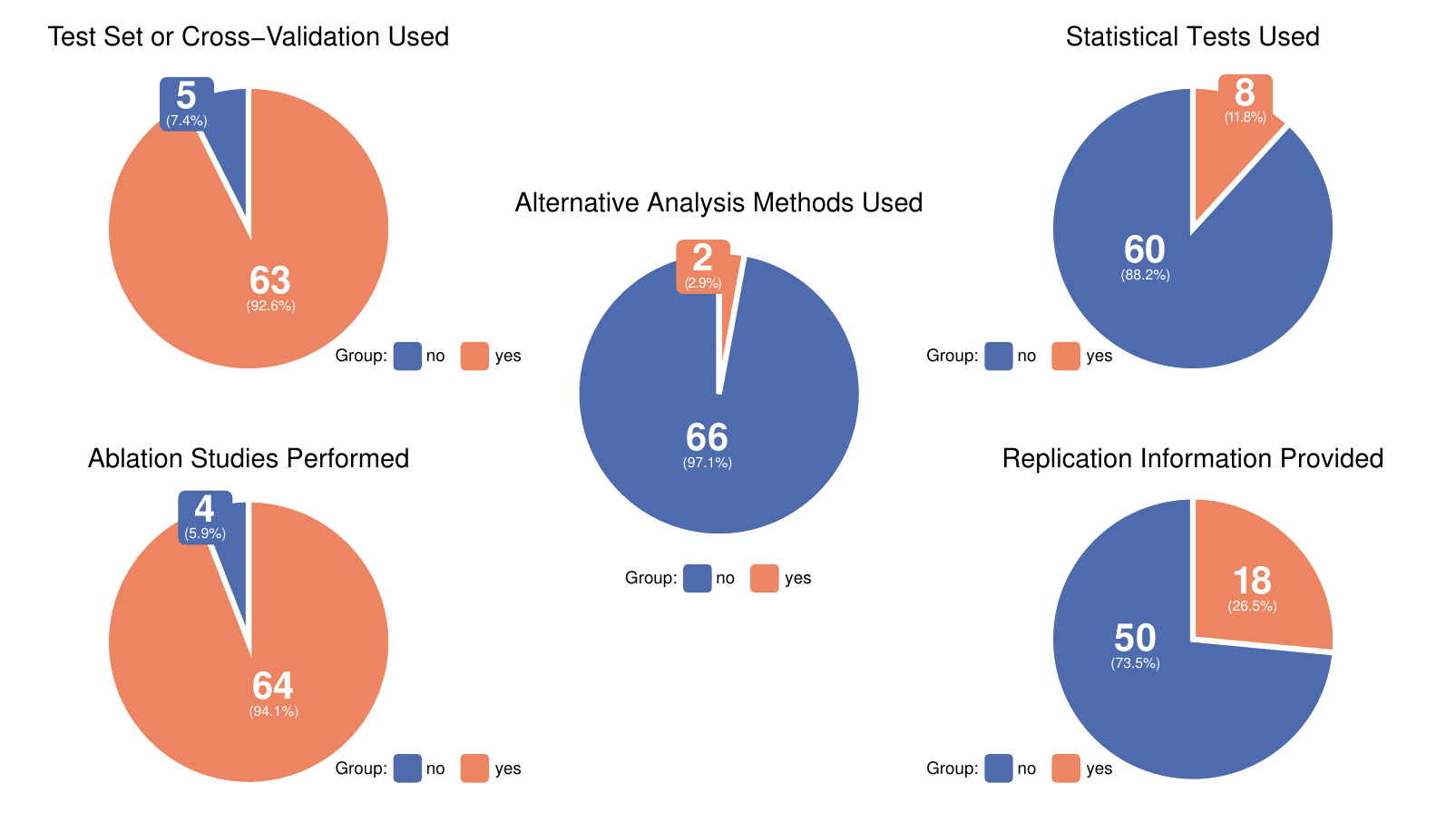}
 \caption{An overview of the validation practices in multimodal experiments. The plots depict use of: a held-out test set or cross-validation (top-left); statistical tests (top-right); alternative analyses like Bayesian methods (centre); ablation studies (bottom-left); and the provision of sufficient information or source code for replication (bottom-right).}
\label{fig:piechartmm}
\end{figure}

\paragraph{Validation procedure}  
The data in Figure~\ref{fig:piechartmm} reveals several notable patterns in research methodology across the studies we analysed. While the vast majority of articles (92.6\%) appropriately utilised test sets for evaluation and conducted ablation studies (94.1\%) to assess component contributions, concerning gaps remain in research rigour. Only 11.8\% of the studies employed statistical tests to validate their findings, and even fewer (2.9\%) incorporated alternative analysis methods such as Bayesian approaches. Most critically, nearly three-quarters (73.5\%) of the articles failed to provide sufficient information or source code for replication. This highlights a considerable transparency issue in the field.

\subsubsection{Extracted data analysis} 
From the literature we reviewed, we identified and extracted performance data from all of the studies that conducted a direct comparison between a multimodal and a unimodal approach. To commence our quantitative synthesis of this data, we computed descriptive statistics for the observed performance differences. These statistics, presented in Table~\ref{tab:mm_summary_metrics_stats}, provide an initial overview of the data's central tendency and dispersion for each metric.

\begin{table}[H]
\centering
\caption{Descriptive statistics for study-level performance differences.}
\centering
\resizebox{\ifdim\width>\linewidth\linewidth\else\width\fi}{!}{
\begin{tabular}[t]{llrrrrrrrr}
\toprule
Metric & Stat. & N & Min & Q1 & Median & Mean & Q3 & Max & SD\\
\midrule
\addlinespace[0.3em]
\multicolumn{10}{l}{\textbf{Accuracy}}\\
\hspace{1em}\cellcolor{gray!10}{Accuracy} & \cellcolor{gray!10}{Mean} & \cellcolor{gray!10}{27} & \cellcolor{gray!10}{-0.33} & \cellcolor{gray!10}{1.81} & \cellcolor{gray!10}{2.70} & \cellcolor{gray!10}{5.62} & \cellcolor{gray!10}{5.86} & \cellcolor{gray!10}{26.87} & \cellcolor{gray!10}{7.03}\\
\hspace{1em}Accuracy & Median & 27 & -1.00 & 1.65 & 2.70 & 5.51 & 5.52 & 28.09 & 7.26\\
\addlinespace[0.3em]
\multicolumn{10}{l}{\textbf{F1-score}}\\
\hspace{1em}\cellcolor{gray!10}{F1-score} & \cellcolor{gray!10}{Mean} & \cellcolor{gray!10}{16} & \cellcolor{gray!10}{0.88} & \cellcolor{gray!10}{1.63} & \cellcolor{gray!10}{3.31} & \cellcolor{gray!10}{3.93} & \cellcolor{gray!10}{4.95} & \cellcolor{gray!10}{9.38} & \cellcolor{gray!10}{2.83}\\
\hspace{1em}F1-score & Median & 16 & 0.75 & 1.44 & 3.35 & 3.91 & 4.91 & 9.55 & 2.89\\
\bottomrule
\end{tabular}}
\label{tab:mm_summary_metrics_stats}
\end{table}
 
\begin{figure}[H]
 \centering
  \includegraphics[scale=0.45,keepaspectratio]{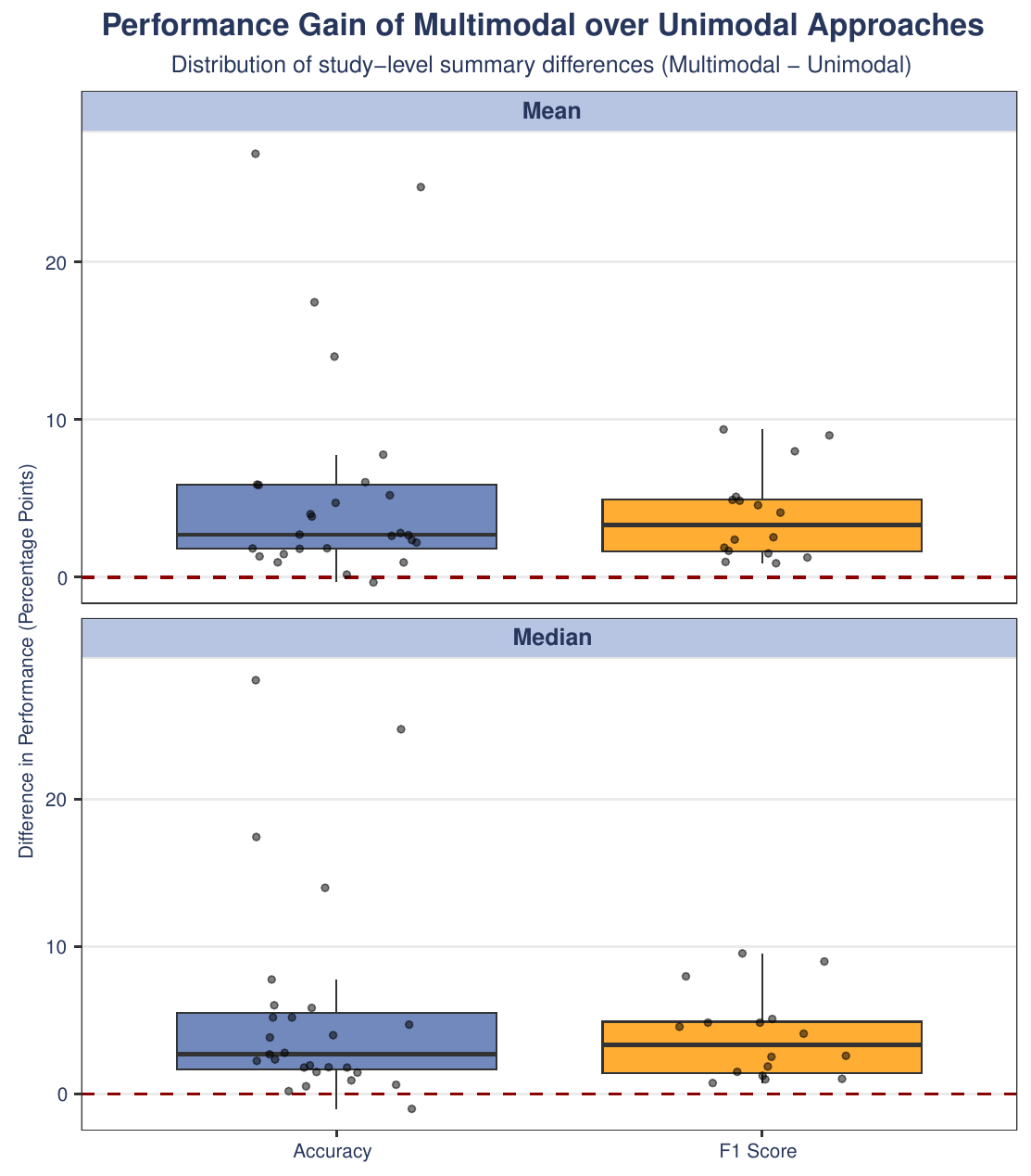}
 \caption{The distribution of performance gains for multimodal over unimodal approaches. The box plots visualise the distribution of study-level performance differences (multimodal--unimodal) across two key metrics. The plots are faceted by the summary statistic used at the study level: mean differences (top panel) and median differences (bottom panel). The horizontal red dashed line at y=0 represents the point of no difference. The black dots depict the individual difference values from each study. The box indicates the interquartile range (IQR), the central line depicts the median of the distribution, and the whiskers extend to 1.5 times the IQR. The number of studies (N) for each metric is presented in Table \ref{tab:mm_summary_metrics_stats}.}
\label{fig:mm_stats_boxplots}
\end{figure}

To better visualise the distributions presented in Table 11, Figure~\ref{fig:mm_stats_boxplots} presents box plots for the mean and median differences across each metric. This visualisation helps to illustrate the spread, potential outliers, and overall skewness of the performance gains.

Having described and visualised the data, we proceeded to formal hypothesis testing with a view to answering our research questions. The results from our comprehensive suite of statistical tests—including the primary meta-analysis, permutation test, Wilcoxon test, and Cliff's delta—are synthesised in Table~\ref{tab:mm_summary_stats_test}. This comprehensive summary enables the direct comparison of findings across different analytical approaches, which highlights the robustness of the results.

\begin{table}[H]
\centering
\caption{A definitive summary of statistical analyses on multimodal performance gains. N: Number of studies. Est. ($\Delta$): Mean difference from meta-analysis. $d_{Cohen}$: Cohen's d effect size. $d_{Cliff}$: Cliff's delta effect size. $r$: Rank-biserial correlation. Significance for respective p-values: *** p < 0.001, ** p < 0.01, * p < 0.05.}
\scalebox{0.75}{
\begin{tabular}[t]{l r >{\bfseries}l c c c c r c c l l}
\toprule
\multicolumn{2}{c}{ } & \multicolumn{3}{c}{Meta-analysis (RQ4)} & \multicolumn{3}{c}{Permutation (RQ1)} & \multicolumn{3}{c}{Wilcoxon (RQ2)} & \multicolumn{1}{c}{Cliff's delta (RQ3)} \\
\cmidrule(l{2pt}r{2pt}){3-5} \cmidrule(l{2pt}r{2pt}){6-8} \cmidrule(l{2pt}r{2pt}){9-11} \cmidrule(l{2pt}r{2pt}){12-12}
Metric & N & Est. $\Delta$ [95\% CI] & p & Sig. & $d_{Cohen}$ & p & Sig. & r [95\% CI] (V) & p & Sig. & $d_{Cliff}$ [95\% CI]\\
\midrule

\rowcolor{gray!6} 
Accuracy & 27 & 5.28 [2.23, 8.32] & 0.0016 & ** & 0.80 & <.001 & *** & 0.99 [0.97, 0.99] (376) & <.001 & *** & 0.31 [-0.01, 0.57] \\

F1-score& 16 & 3.29 [-0.0237, 6.6] & 0.0511 &    & 1.39 & <.001 & *** & 1 [1, 1] (136) & <.001 & *** & 0.27 [-0.15, 0.60] \\
\bottomrule
\end{tabular}
}
\label{tab:mm_summary_stats_test}
\end{table}

\subsubsection{Summary}
Our two-stage analysis delivers a robust narrative on the effectiveness of multimodal approaches, revealing a clear benefit for accuracy and a more complex, model-dependent picture for F1-score. The key findings from each stage are as follows: (1) \textit{Exploratory analysis}: a strong signal with underlying uncertainty. Our initial nonparametric tests established a strong and statistically significant positive effect of multimodality. Both the permutation test (which evaluates the mean) and the Wilcoxon signed-rank test (which evaluates the median) revealed a highly significant advantage for accuracy and F1-score ($p < .001$ for all). This confirmed that the positive finding was a typical result and was not driven by outliers. However, an assessment of the practical magnitude using Cliff's delta introduced critical nuance. While indicating a positive effect, the wide confidence intervals for both metrics crossed zero, which highlights substantial uncertainty around the effect's true consistency and magnitude. This underscored the need for a formal meta-analysis to derive a more stable, weighted estimate; and (2) \textit{Primary meta-analysis}: a clear benefit for accuracy, but a fragile finding for F1-score. The primary random effects meta-analysis aimed to synthesise the findings into a precise estimate, but the results differed starkly by metric: \textit{For accuracy}, the meta-analysis confirmed a clear and significant benefit, yielding a robust average improvement of +5.28 percentage points (95\% CI [2.23, 8.32]; $p = 0.0016$). This finding was accompanied by high heterogeneity ($I^2 = 82.6\%$), reflecting substantial variability across study contexts; and \textit{In stark contrast}, the analysis for F1-score revealed the fragility of the evidence. Our primary weighted model found a nonsignificant improvement of +3.29 points (95\% CI [-0.02, 6.60]; $p = 0.0511$) with notably low heterogeneity ($I^2 \approx 10\%$), which directly contradicts the significant findings from the unweighted exploratory tests. This model dependence highlights that while the trend for F1-score is positive, the evidence base is too methodologically fragile to support a single, definitive conclusion.

Synthesising these findings, we conclude \textbf{that the evidence for multimodality's benefit differs significantly by the performance metric used. For \textbf{accuracy}, the narrative is clear and consistent: we found a genuine, though highly variable, performance gain across a broad set of research contexts. In contrast, the \textbf{F1-score} result represents a tentative positive trend from a smaller, methodologically compromised dataset. Its marginal, nonsignificant finding in the primary analysis reflects a high degree of sensitivity to statistical artifacts introduced by poor data reporting}. This highlights a critical challenge for meta-research in machine learning: final conclusions depend not only on what is published, but critically on the quality of the statistical reporting in those publications.

\subsection{Multiview representation approaches}
\subsubsection{Evaluation details}
\begin{figure}[H]
 \centering
  \includegraphics[scale=0.45,keepaspectratio]{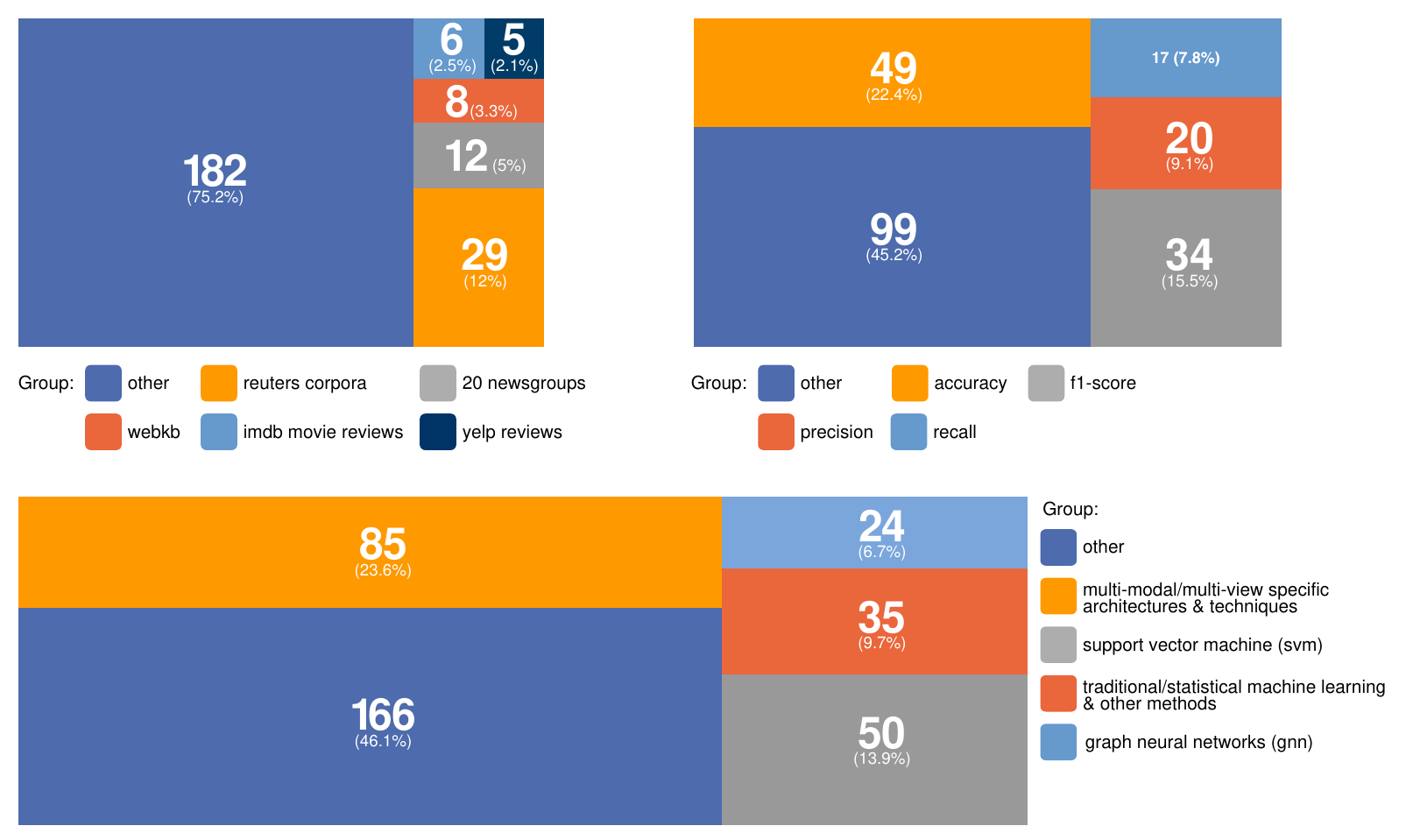}
 \caption{An overview of the evaluation practices in multiview experiments: The top-left plot depicts the frequency distribution of the datasets; the top-right plot presents the frequency distribution of the evaluation metrics; and the bottom plot displays the distribution of the baseline models used for comparison.}
\label{fig:mvdsmetmod}
\end{figure}

\paragraph{Datasets, models selected for comparison, and performance metrics} 
The analysis of information-fusion-based document classification literature reveals several noteworthy patterns. In Figure~\ref{fig:mvdsmetmod}, we observe that in terms of evaluation methodologies, researchers predominantly employ diverse, specialised datasets (75.2\%), while standard benchmarks like the Reuters corpora (12\%) and 20 Newsgroups (5\%) maintain relevance but are less dominant. Regarding baseline comparison models, a substantial portion (46.1\%) falls outside of common categories, which reflects the field's methodological diversity. Multimodal/multiview-specific architectures represent a significant comparison point (23.6\%), with traditional approaches like SVM (13.9\%) continuing to be relevant. For performance evaluation, while numerous custom metrics are employed (45.2\%), researchers continue to rely heavily on standard metrics with accuracy (22.4\%) and F1-score (15.5\%) being the most prevalent. This distribution pattern across all three dimensions—datasets, comparison models, and metrics—demonstrates both the field's methodological diversity and its adherence to certain established benchmarks and evaluation practices.

\begin{figure}[H]
 \centering
  \includegraphics[scale=0.45,keepaspectratio]{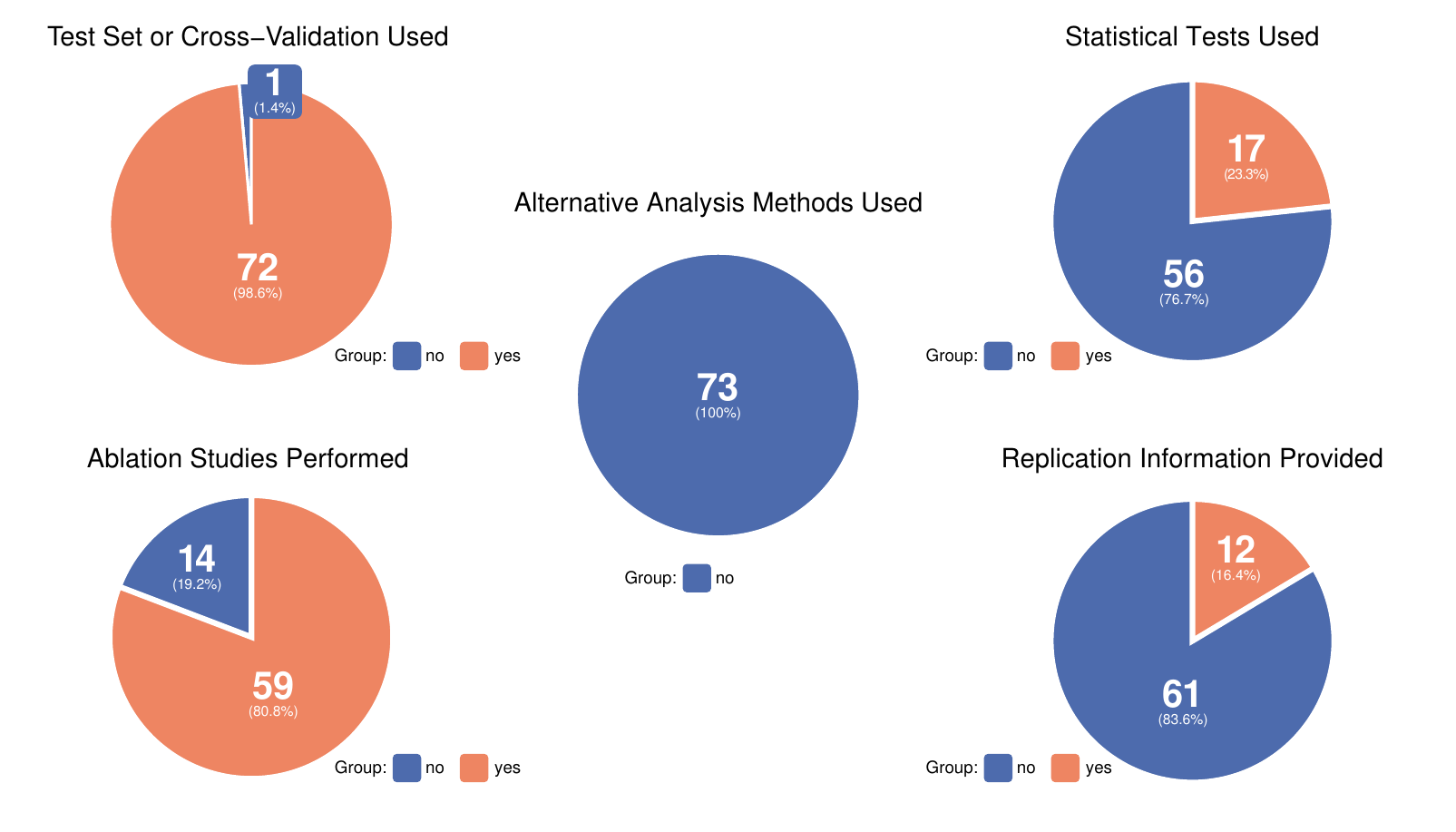}
 \caption{An overview of the validation practices in multiview experiments. The plots depict use of: a held-out test set or cross-validation (top-left); statistical tests (top-right); alternative analyses like Bayesian methods (centre); ablation studies (bottom-left); and the provision of sufficient information or source code for replication (bottom-right).}
\label{fig:piechartmv}
\end{figure}

\paragraph{Validation procedure} 
Our analysis of 73 articles reveals several notable trends in information-fusion-based document classification research (Figure~\ref{fig:piechartmv}). While nearly all studies (98.6\%) appropriately employed test sets or cross-validation procedures for evaluation, and a substantial majority (80.8\%) conducted ablation studies to analyse component contributions, sizeable methodological gaps persist. Strikingly, statistical significance testing was performed in only 23.3\% of the articles, and none employed alternative analytical approaches such as Bayesian methods. Perhaps most concerning is the replication challenges in reproducibility that are evident in the field: only 16.4\% of the articles provided sufficient information or code to enable reproduction of their results. These findings highlight a strong focus on performance evaluation, but reveal substantial shortcomings in statistical validation and research reproducibility. This suggests the existence of opportunities for improving methodological rigour in future work.

\subsubsection{Extracted data analysis} 
From the literature we reviewed, we identified and extracted performance data from each of the studies that conducted a direct comparison between a multiview and a singleview approach. To commence our quantitative synthesis of that data, we computed descriptive statistics for the observed performance differences. Those statistics, summarised in Table~\ref{tab:mv_summary_metrics_stats}, provide an initial overview of the data's central tendency and dispersion for each metric.

\begin{table}[H]
\centering
\caption{Descriptive statistics for study-level performance differences.}
\centering
\resizebox{\ifdim\width>\linewidth\linewidth\else\width\fi}{!}{
\begin{tabular}[t]{llrrrrrrrr}
\toprule
Metric & Stat. & N & Min & Q1 & Median & Mean & Q3 & Max & SD\\
\midrule
\addlinespace[0.3em]
\multicolumn{10}{l}{\textbf{Accuracy}}\\
\hspace{1em}\cellcolor{gray!10}{Accuracy} & \cellcolor{gray!10}{Mean} & \cellcolor{gray!10}{20} & \cellcolor{gray!10}{-0.20} & \cellcolor{gray!10}{1.46} & \cellcolor{gray!10}{2.14} & \cellcolor{gray!10}{4.34} & \cellcolor{gray!10}{5.29} & \cellcolor{gray!10}{26.99} & \cellcolor{gray!10}{5.99}\\
\hspace{1em}Accuracy & Median & 20 & -0.20 & 1.26 & 2.35 & 4.28 & 5.29 & 26.99 & 5.98\\
\addlinespace[0.3em]
\multicolumn{10}{l}{\textbf{F1-score}}\\
\hspace{1em}\cellcolor{gray!10}{F1 Score} & \cellcolor{gray!10}{Mean} & \cellcolor{gray!10}{17} & \cellcolor{gray!10}{-6.15} & \cellcolor{gray!10}{1.92} & \cellcolor{gray!10}{2.24} & \cellcolor{gray!10}{3.28} & \cellcolor{gray!10}{4.60} & \cellcolor{gray!10}{11.47} & \cellcolor{gray!10}{3.68}\\
\hspace{1em}F1 Score & Median & 17 & -6.15 & 2.00 & 2.64 & 3.20 & 4.50 & 12.20 & 3.75\\
\addlinespace[0.3em]
\multicolumn{10}{l}{\textbf{Precision}}\\
\hspace{1em}\cellcolor{gray!10}{Precision} & \cellcolor{gray!10}{Mean} & \cellcolor{gray!10}{5} & \cellcolor{gray!10}{-0.57} & \cellcolor{gray!10}{1.72} & \cellcolor{gray!10}{2.25} & \cellcolor{gray!10}{2.68} & \cellcolor{gray!10}{4.09} & \cellcolor{gray!10}{5.90} & \cellcolor{gray!10}{2.45}\\
\hspace{1em}Precision & Median & 5 & -1.59 & 1.52 & 2.25 & 2.38 & 3.84 & 5.90 & 2.79\\
\addlinespace[0.3em]
\multicolumn{10}{l}{\textbf{Recall}}\\
\hspace{1em}\cellcolor{gray!10}{Recall} & \cellcolor{gray!10}{Mean} & \cellcolor{gray!10}{6} & \cellcolor{gray!10}{0.40} & \cellcolor{gray!10}{2.15} & \cellcolor{gray!10}{4.35} & \cellcolor{gray!10}{5.55} & \cellcolor{gray!10}{6.24} & \cellcolor{gray!10}{15.89} & \cellcolor{gray!10}{5.54}\\
\hspace{1em}Recall & Median & 6 & 0.40 & 2.15 & 4.35 & 5.49 & 6.00 & 15.89 & 5.54\\
\bottomrule
\end{tabular}}
\label{tab:mv_summary_metrics_stats}
\end{table}
 
\begin{figure}[H]
 \centering
  \includegraphics[scale=0.35,keepaspectratio]{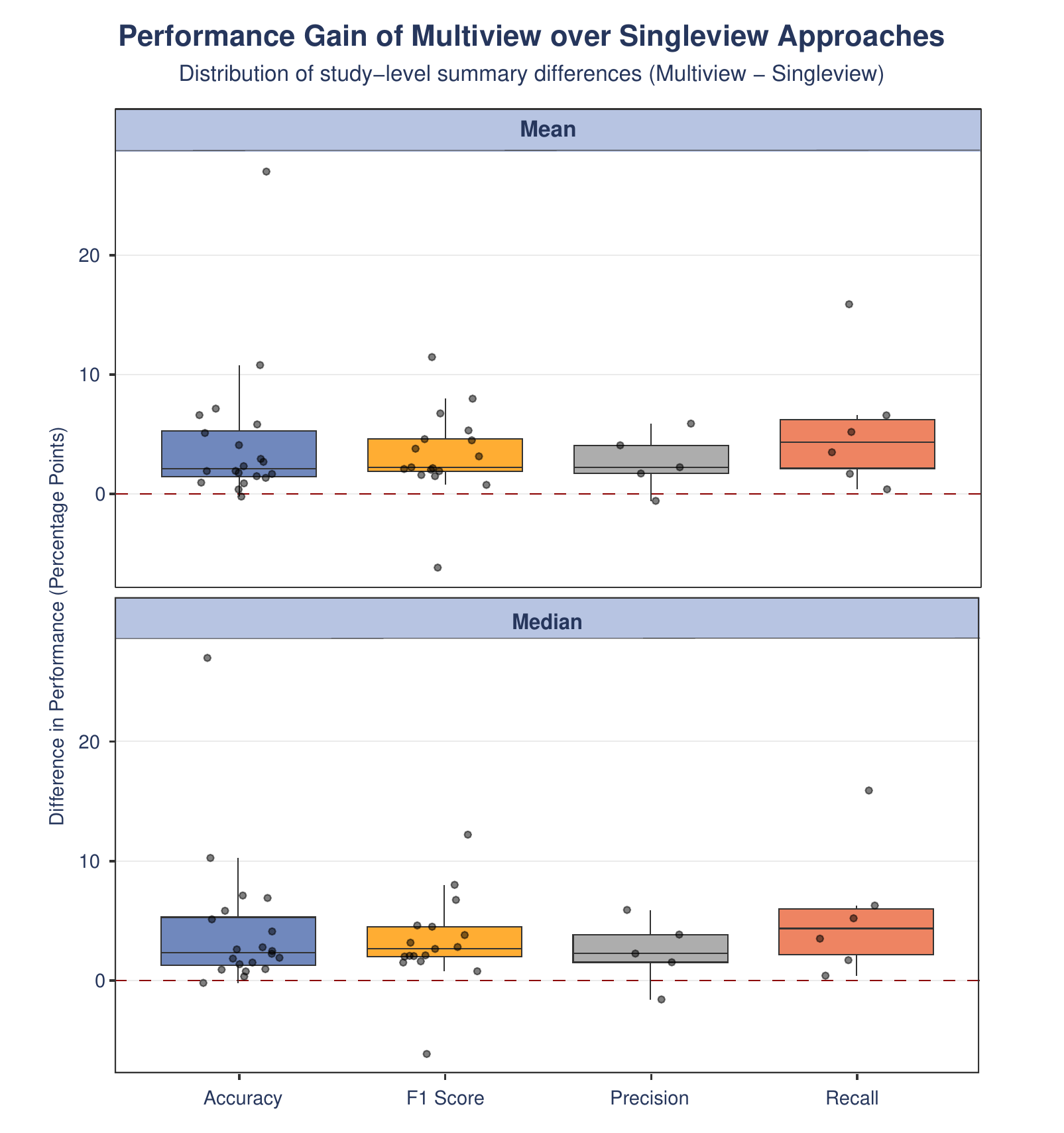}
 \caption{The distribution of performance gains for multiview over singleview approaches. The box plots visualise the distribution of study-level performance differences (multiview--singleview) across four key metrics. The plots are faceted by the summary statistic used at the study level: mean differences (top panel) and median differences (bottom panel). The horizontal red dashed line at y=0 represents the point of no difference. The black dots depict the individual difference value from each study. The box indicates the interquartile range (IQR), the central line depicts the median of the distribution, and the whiskers extend to 1.5 times the IQR. The number of studies (N) for each metric is provided in Table \ref{tab:mv_summary_metrics_stats}.}
\label{fig:mv_stats_boxplots}
\end{figure}

To better visualise the distributions detailed in the table, Figure~\ref{fig:mv_stats_boxplots} presents box plots for the mean and median differences across each metric. This visualisation helps to illustrate the spread, potential outliers, and overall skewness of the performance gains.

Having described and visualised the data, we proceeded to formal hypothesis testing with a view to answering our research questions. The results from our comprehensive suite of statistical tests—including the primary meta-analysis, permutation test, Wilcoxon test, and Cliff's delta—are synthesised in Table~\ref{tab:mv_summary_stats_test}. This comprehensive summary enables a direct comparison of findings across different analytical approaches. This highlights the robustness of the results.

\begin{table}[H]
\centering
\caption{A definitive summary of statistical analyses on multiview performance gains. N: Number of studies. Est. ($\Delta$): Mean difference from meta-analysis. $d_{Cohen}$: Cohen's d effect size. $d_{Cliff}$: Cliff's delta effect size. $r$: Rank-biserial correlation. Significance for respective p-values: *** p < 0.001, ** p < 0.01, * p < 0.05.}
\scalebox{0.75}{
\begin{tabular}[t]{l r >{\bfseries}l c c c c r c c l l}
\toprule
\multicolumn{2}{c}{ } & \multicolumn{3}{c}{Meta-analysis (RQ4)} & \multicolumn{3}{c}{Permutation (RQ1)} & \multicolumn{3}{c}{Wilcoxon (RQ2)} & \multicolumn{1}{c}{Cliff's delta (RQ3)} \\
\cmidrule(l{2pt}r{2pt}){3-5} \cmidrule(l{2pt}r{2pt}){6-8} \cmidrule(l{2pt}r{2pt}){9-11} \cmidrule(l{2pt}r{2pt}){12-12}
Metric & N & Est. $\Delta$ [95\% CI] & p & Sig. & $d_{Cohen}$: & p & Sig. & r [95\% CI] (V) & p & Sig. & $d_{Cliff}$ [95\% CI]\\
\midrule

\rowcolor{gray!6} 
    Accuracy & 
    20 & 
    4.67 [0.37, 8.98] & 
    0.035 & 
    * & 
    0.725 & 
    <.001 &
    *** &    
    0.99 [0.97, 1.00] (209) & 
    <.001 & 
    *** &     
    0.22 [-0.15, 0.54]
    \\

    F1-score& 
    17 & 
    3.08 [0.15, 6.02] & 
    0.044 & 
    * & 
    0.891 & 
    0.002 & 
    ** &    
    0.82 [0.54, 0.93] (139) & 
    0.002 & 
    ** & 
    0.16 [-0.24, 0.52]
    \\

    \rowcolor{gray!6} 
    Precision & 
    5 & 
    1.16 [-7.11, 9.43] & 
    0.591 &
    &
    1.093 & 
    0.129 &
    &    
    0.87 [0.34, 0.98] (14) &     
    0.125 &  
    &     
    0.12 [-0.61, 0.74]
    \\

    Recall & 
    6 & 
    5.87 [1.55, 10.18] & 
    0.020 & 
    * & 
    1.000 & 
    0.032 & 
    * &     
     1.00 [1.00, 1.00] (21) & 
    0.031 & 
    * &    
    0.22 [-0.47, 0.74]
    \\
\bottomrule
\end{tabular}
}
\label{tab:mv_summary_stats_test}
\end{table}

\subsubsection{Summary}
Our two-stage analysis provides a robust and coherent narrative on the effectiveness of multiview approaches. The key findings from each stage are as follows: (1) \textit{Exploratory analysis: a consistent but modest signal}. Our initial nonparametric tests established the presence of a consistent positive effect. Both the permutation test (which evaluates the mean) and the Wilcoxon signed-rank test (which evaluates the median) revealed a statistically significant advantage for multiview approaches in accuracy, F1-score, and recall ($p < .05$ for all), but not for precision. This confirmed that the finding was a typical result in the literature and was not driven by outliers. However, an assessment of the practical magnitude using Cliff's delta indicated that this statistically significant effect was of a `small' size. The wide confidence intervals from this analysis also highlighted significant uncertainty, which underscores the need for a more powerful, weighted analysis; and (2) \textit{Primary meta-analysis: a precise estimate of the effect}. The primary random effects meta-analysis confirmed the conclusions of the exploratory stage and provided the best quantitative estimates of the effects. We found a statistically significant average improvement for accuracy (+4.67 percentage points; 95\% CI [0.37, 8.98]; $p = 0.035$), F1-score (+3.08 points; 95\% CI [0.15, 6.02]; $p = 0.044$), and recall (+5.87 points; 95\% CI [1.55, 10.18]; $p = 0.020$). Heterogeneity was high for accuracy ($I^2 = 90.2\%$) but negligible for F1-score and recall ($I^2 \approx 0\%$), likely due to variance imputation in the latter metrics. As in the initial analysis, the effect on precision remained nonsignificant.

Taken together, our two-stage analytical approach delivers strong, convergent evidence. The initial exploratory tests established the presence and robustness of a positive signal, while the primary meta-analysis provided the best quantitative estimate of its size. \textbf{We conclude that strong evidence exists for a consistent, statistically significant, but practically modest advantage for multiview approaches across the accuracy, F1-score, and recall metrics.}

\subsection{Multimodal and multiview approaches summary}
This section synthesises the results of our comprehensive two-stage analysis of multimodal and multiview document classification studies. Across both representation strategies, we found converging evidence of performance benefits compared to unimodal or singleview baselines, although the strength and reliability of those effects varied by metric and methodological rigour. (1) \textit{For multimodal approaches}, the evidence was mixed and highly dependent on the evaluation metric. A clear and statistically significant benefit was observed for accuracy, with a random effects meta-analysis confirming an average gain of +5.28 percentage points (95\% CI [2.23, 8.32]; $p = 0.0016$). In contrast, the effect on F1-score proved statistically fragile. While initial nonparametric tests suggested a significant advantage, the primary meta-analytic estimate narrowly fell short of significance ($p = 0.0511$). This inconsistency highlights the sensitivity of findings to methodological limitations---particularly to poor statistical reporting---which leaves the evidence for metrics beyond accuracy positive but ultimately inconclusive; (2) By contrast, \textit{multiview approaches} demonstrated more consistent benefits. We observed statistically significant improvements for accuracy, F1-score, and recall across both exploratory and meta-analytic phases. The weighted meta-analysis confirmed modest but robust average gains—accuracy (+4.67 points), F1-score (+3.08 points), and recall (+5.87 points)—with all $p$-values below the 0.05 threshold. Precision was the only metric that exhibited no significant difference. Although the practical magnitude of these effects was small (as indicated by Cliff’s delta), their statistical robustness across diverse experimental contexts enhances their credibility; and (3) \textit{Despite these positive findings, both strands of the literature suffer from persistent methodological shortcomings}. Most notably, formal statistical testing and research reproducibility remain rare, with fewer than a quarter of the studies including such analyses or providing the code and data necessary for replication. This systemic issue limits the interpretability and generalisability of many reported gains.

In summary, our meta-analysis indicates that both multimodal and multiview strategies offer a genuine advantage over their unimodal and singleview counterparts, particularly for the accuracy metric. The evidence base for multiview approaches proved more consistently positive, demonstrating significant gains across a wider range of metrics. Nevertheless, the practical impact remains modest, and the field would benefit greatly from improved methodological transparency and statistical validation.

\section{Discussion}\label{sec:discussion}

\subsection{A unified framework for information fusion}
One of the most valuable outcomes of this review is the development of a conceptual (or formal) framework that unifies and organises key ideas in document classification, revealing how they interrelate and suggesting how they might be extended or tested empirically. This framework casts each component of a document-classification pipeline—tokenisation, feature extraction, encoding, fusion, recognition—in a consistent notation: $\text{Representation: } R = (F,E,M)$, $\text{Pattern: } P = (S,C,T)$, and $\text{Model: } \mathcal{M} = (P,R,RR)$. By doing so, it provides: (1) \textit{A clear taxonomy of concepts}: It becomes straightforward to discuss and combine multiple data sources (or multiple views on the same data) within a single, uniform structure; (2) \textit{Guidance for practical implementation}: Researchers can identify the stage at which they want to integrate a new data source or technique, using explicit fusion strategies at the feature or decision level; (3) \textit{Clarity in comparing approaches}: The question of \emph{how} and \emph{when} data is fused can be addressed consistently. Researchers can evaluate differences in published methods by asking whether they fuse at \((F)\) or \((E)\), or whether they integrate classifiers using different \(RR\)s; and (4) \textit{Potential for extensibility}: The framework is intentionally modular. As research evolves (e.g. adoption of LLMs, real-time data streaming), these new elements can  `plug in' without disrupting the conceptual architecture.

\subsection{Alignment with formalisms in computational science}
From a broader perspective, the framework aligns well with similar formal approaches in pattern recognition and machine learning. In many fields, structured theoretical models offer both a holistic overview (i.e. `zooming out' on the entire pipeline) and granular detail (`zooming in' on particular transitions between stages). Here, a parallel could be drawn to algebraic classification theory or other systematic methods in engineering and computational sciences, in which abstract conceptual tools are used to describe and unify diverse methods under a single theoretical umbrella. By defining precisely how features, encodings, and meanings relate, researchers obtain a robust lens for analysing existing techniques and adapting them to novel contexts.

\subsection{Synthesis of extracted results and implications for research and practice}

\subsubsection{Multimodal models}
The primary meta-analysis reveals a clear, statistically significant advantage for multimodal approaches in terms of accuracy, with a mean improvement of +5.28 percentage points ($p = 0.0016$).  This contrasts sharply with the findings for F1-score, in which the effect was marginal and highly dependent on the statistical model used (+3.29 points, $p = 0.0511$). This quantitative divergence points to a more profound qualitative story: the success of multimodal fusion is not determined by a single superior technique, but by the fundamental nature of the interaction between modalities—whether they provide synergistic reinforcement or create a strategic trade-off.

Our analysis of high-impact studies uncovers three primary patterns. The most common is `synergistic fusion', in which the multimodal result outperforms any single modality. This pattern itself has two distinct drivers. In the work of~\cite{liangz2023} on fake news detection, synergy arises from the fusion of complementary information; the text and image provide different, contextually-related clues that, when combined, lead to a more accurate classification than either could achieve alone. A second form of synergy is demonstrated in~\cite{braz2020}, in which an artificially generated text modality is fused with visual data. The study demonstrates that this fusion boosts accuracy, even when the generated text is of inconsistent quality, and can provide complementary information that corrects errors made by the visual-only model. The final, and most nuanced, pattern is the `less is more' principle revealed by the analysis of~\cite{kenny2023} in emotion recognition. The authors found that a model that combines the two best modalities (text and motion capture) outperformed a model that fused all three available modalities. This demonstrates that the simple addition of more data streams does not guarantee better performance and that the strategic selection of modalities is crucial. Together, these cases demonstrate that the most effective multimodal systems are those in which the fusion strategy is aligned deliberately with both the relationship between the modalities and the ultimate performance goals of the task.

These findings carry major implications for both practitioners and researchers in the multimodal domain. (1) \textbf{For practitioners}, the central finding is that the selection of fusion architecture is a strategic decision dictated by the task's underlying semantic relationship. For tasks that involve synergistic modalities in which signals are complementary (e.g. emotion recognition), simple fusion methods are highly effective. For tasks that require the detection of semantic incongruity (e.g. fake news detection), more sophisticated attention-based mechanisms are necessary. This selection is predicated on the foundational use of powerful, pretrained unimodal backbones, in which the fusion layer's primary role is to act as a `smart bridge' between them; and (2) \textbf{For researchers}, our analysis suggests a shift in focus from the mere proposal of novel fusion algorithms towards more foundational challenges. The next research frontier lies in: (1) developing a formal taxonomy of intermodal relationships (e.g. synergy, redundancy, incongruity) to guide architectural choice \emph{a priori}; (2) exploring creative data-centric strategies, such as the artificial synthesis of modalities, to unlock performance in traditionally unimodal domains (the work of \cite{kenny2023} provides a compelling case study, which demonstrates that while targeted data augmentation significantly improved an underperforming third modality, the model with the two best-selected modalities still yielded the highest overall performance, underscoring the critical interplay between modality selection and data quality); and (3) designing architectures that effectively integrate specialised, pretrained vision--language models as distinct components. To support this trajectory, we issue a call for more rigorous reporting standards, urging authors to include unimodal baselines, F1-scores for imbalanced datasets, and full ablation studies to facilitate robust meta-analysis and to build a cumulative science. 

\subsubsection{Multiview models}
The primary meta-analysis reveals a consistent, statistically significant, but practically modest advantage for multiview approaches, improving mean accuracy by 4.67 percentage points (95\% CI [0.37, 8.98]) and F1-score by 3.08 points (95\% CI [0.15, 6.02]). This quantitative result, however, belies a more profound qualitative story: the magnitude of success is not determined by a single superior fusion strategy (e.g. early vs. late), but by the deliberate alignment of the fusion method with the problem context and available learning paradigms.

Our analysis of high-impact studies uncovers two primary drivers of exceptional performance. First, the task context is paramount, with the most substantial gains observed when views are naturally complementary or `orthogonal', such as in multilingual categorisation in which different languages provide distinct feature spaces. The work of~\cite{aminim2009} further demonstrates that when such views are unavailable, they can be created artificially via techniques like machine translation, transforming data augmentation into a powerful fusion enabler. Second, a clear paradigm shift is observable. While classic late-fusion ensembles remain highly effective, the most dramatic modern gains, as seen in~\cite{bhatt2019}~\footnote{The authors present a multimodal architecture that can also be used for multiview analysis, and it is for this case, i.e. multiview classification of text documents, that the results are presented.}, are driven by early-fusion architectures that leverage transfer learning from large, pretrained models (e.g. ResNet, Word2Vec).

These findings carry crucial implications for both practitioners and researchers. (1) \textbf{For practitioners}, the choice between early and late fusion should be guided by the task's particular needs, not by dogma. Late fusion offers a robust and simple path when strong, independent classifiers can be trained for each view. In contrast, early fusion is the route to state-of-the-art performance when leveraging the rich representations from modern pretrained feature extractors. The key takeaway is to prioritise view complementarity; if the views' errors are uncorrelated, fusion is highly likely to succeed; and (2) \textbf{For researchers}, our analysis suggests a shift in focus. Rather than the invention of new fusion algorithms, the next research frontier lies in: (1) the development of theoretical frameworks to quantify view complementarity \emph{a priori}, (2) the exploration of novel methods for the artificial creation of high-quality views from single-source data, and (3) the design of efficient fusion architectures specifically for integrating representations from large foundational models. This heralds a paradigm shift in which the core task moves from end-to-end learning to the sophisticated combination of powerful, precomputed semantic spaces. To enable such progress, we also issue a call for greater methodological rigour in reporting, urging authors to include singleview baselines and standard deviations to facilitate future synthesis and to build a more robust, cumulative science.

\subsection{Limitations}
While this systematic review provides a comprehensive framework and the first quantitative synthesis in its field, certain limitations should be acknowledged: (1) \textit{Granularity of quantitative synthesis}: Our meta-analysis successfully estimates the overall performance advantage of multimodal and multiview approaches. However, the diversity of methods and inconsistent reporting in the primary literature prevented more granular sub-group analyses. Consequently, we could not statistically compare the relative effectiveness of particular fusion strategies (e.g. early vs. late fusion) nor quantify how performance gains vary across different modality pairings (e.g. text--image vs. text--audio); (2) \textit{Bias in the primary literature}: Our review accurately reflects the current state of the field, which is  dominated by text--image and text--text fusion. While we discuss other modalities, the relative scarcity of research on combinations that involve audio, time-series, or sensor data means that our conclusions are most robustly supported for the more common pairings; (3) \textit{Inability to control for data quality in meta-analysis}: Although our qualitative analysis explicitly identifies poor data quality (e.g. OCR errors, label noise) as a critical challenge, our quantitative meta-analysis could not control for these unobserved variables. The effect sizes are synthesised from studies with varying and often unreported levels of data quality. This introduces a source of statistical heterogeneity that, while modelled, cannot be fully partitioned and explained; and (4) \textit{Search scope limitation}: Our systematic search was conducted using a single database (Scopus) to ensure query transparency and reproducibility. While this deliberate choice enhances methodological rigour, it introduces a potential selection bias by excluding studies not indexed in Scopus (Our Scopus search improves transparency/reproducibility, and we treat completeness conservatively). Although our findings represent a robust synthesis of a large and representative corpus, their generalisability should be considered with this explicit methodological boundary in mind.    

\section{Conclusion}\label{sec:conclusion}
This systematic review concludes that while information fusion offers a demonstrable performance advantage in document classification, its practical benefits are nuanced and its true potential is currently constrained by wide methodological gaps in the literature. In response to our research questions, this work establishes a unifying formal framework that anchors modern representation learning in classical information fusion theory, provides the first quantitative meta-analytic estimate of fusion's effectiveness, and offers evidence-based recommendations. Our primary contributions are summarized in Table~\ref{tab:conclusion_summary}.

\begin{table}[H] 
\centering 
\caption{Key conclusions and their implications.}\vspace{-10pt}
\label{tab:conclusion_summary} 
\small 
\renewcommand{\arraystretch}{1.3} 
\begin{tabularx}{\textwidth}{@{} >{}p{0.22\textwidth} >{\RaggedRight\arraybackslash}X @{}} \toprule \textbf{Key conclusion} & \textbf{Implication} \\ 

\midrule 

A unifying formal framework grounded in Information Fusion theory & 
The proposed frameworks for representation ($R$), pattern ($P$), and model ($M$) act as a theoretical bridge. They enable researchers to map ad-hoc machine learning architectures (e.g., attention, ensembles) onto rigorous fusion paradigms—such as Bayesian opinion pools and Dempster--Shafer evidential reasoning—facilitating systematic comparison and theoretical validation. \\ 

\addlinespace

Quantified, but nuanced performance gains & 
Fusion is not a panacea. \textbf{Multimodal} approaches yield a clear benefit for accuracy (+5.28\%, $p=.0016$) but a fragile one for F1-score. \textbf{Multiview} approaches demonstrate more consistent, albeit modest, gains across accuracy, F1-score, and recall. The selection of a fusion strategy must be context-dependent. \\

\addlinespace
    
Challenges in reproducibility & 
The validity of many reported gains is questionable. With only 11.8\% (multimodal) and 23.3\% (multiview) of the studies employing statistical tests, and even fewer sharing replication materials, the field suffers from a lack of transparency and statistical validation that hinders cumulative progress. \\

\addlinespace
    
Strategic guidance over algorithmic superiority &
The most successful applications do not hinge on a single `best' algorithm, but on the strategic alignment of the fusion method with the task context (e.g. using complementary vs. redundant information) and the thoughtful integration of powerful pretrained models. \\

\bottomrule

\end{tabularx}
\end{table}

Our qualitative and quantitative analyses converge on a critical insight: the empirical evidence for information fusion, while positive, is less robust than the frequency of its application would suggest. The fragility of the F1-score result in the multimodal meta-analysis, for example, is a direct consequence of poor statistical reporting and the lack of standardized benchmarks identified in the qualitative review. Future progress depends less on inventing incrementally novel fusion algorithms and more on addressing foundational challenges.

Therefore, we conclude with two primary recommendations: (1) \textbf{For practitioners}, the central lesson is to prioritize  strategy over complexity. This means rejecting the flawed assumption of a `presumed algorithmic hierarchy'—in which complex methods like cross-modal transformers are automatically considered superior to simpler strategies like feature concatenation or decision voting—and instead making deliberate choices based on the problem's unique structure. This requires practitioners to: (1.1) \textit{Prioritize view diversity}: Seek out or even artificially create complementary or uncorrelated views (e.g. via machine translation), as these provide the largest informational gain. The power of modern transformers, for instance, can be partly understood as a dynamic, learned fusion of multiple `views' of the input sequence via the self-attention mechanism; (1.2) \textit{Match architecture to the task}: The fusion architecture must be matched to the task's semantic nature. If a task can be decomposed into a set of reinforcing, complementary views, a simple and efficient fusion strategy is often sufficient. In contrast, for tasks that hinge on the resolution of semantic incongruity—such as detecting sarcasm from conflicting textual and auditory cues—more sophisticated, attention-based architectures are necessary to model these complex interactions; and (1.3) \textit{Justify every modality}: Each added modality or view must be justified rigorously. A careful cost--benefit analysis, supported by ablation studies, is essential to confirm that its performance contribution outweighs the inevitable increase in computational cost, training time, and overall model complexity; and (2) \textbf{For researchers}, these practical guidelines point to a necessary shift in the scientific agenda: from an algorithm-driven science to a methodology- and theory-driven one. 

To advance the science of document information fusion, research should focus on: (i) \textit{Quantifying view interactions via fusion theory}: Developing formal measures to characterize inter-view conflict and complementarity \textit{a priori} (e.g., using entropy-based or evidential distance metrics from Dempster--Shafer theory) to predict fusion success before training; (ii) \textit{Formalizing uncertainty propagation}: Investigating how uncertainty in learned representations (e.g., from deep encoders) can be propagated through the fusion stage using Bayesian or belief-function formalisms, rather than relying solely on point-estimate concatenations; and (iii) \textit{Methodological rigour}: Adopting a community-wide commitment to rigorous statistical validation and reproducibility, to distinguish genuine gains derived from fusion from those resulting from stochastic noise.

Ultimately, this review serves as both a benchmark and a theoretical roadmap. By formally linking document classification to information fusion paradigms, it provides the necessary tools to build a more robust, transparent, and cumulative science of fusion in text-centric systems.

\appendix
\section{Supplemental items}\label{app:supplement}
All supplementary materials, replication data, and code are archived on Zenodo:
\url{https://doi.org/10.5281/zenodo.17141560}. The record includes: (1) \
Supplementary Materials (PDF) — documents the database search queries, describes the datasets and codebase, and provides extended summaries complementing the main text; (2) Datasets - extracted data used in the quantitative meta-analysis; (3) R scripts - to reproduce the statistical analyses and figures; and (4) Technical reports (HTML) — two reports generated from the scripts with methodological details, tables, and results.

\noindent\emph{Review-only link.} A temporary, non-persistent preview URL may be provided to the editors during peer review; it will be replaced by the DOI in the final version.

\section{Declaration of generative AI and AI-assisted technologies in the writing process}
During the preparation of this work, the author(s) used openrouter.ai and language model such as GPT, Grok 4 and Gemini with reasoning mode in order to improve readability and to maintain consistent scientific language. After using this tool/service, the author(s) reviewed and edited the content as needed and take(s) full responsibility for the content of the published article.

\begingroup
\setstretch{1.0}            
\setlength{\bibsep}{1pt}    
\setlength{\bibhang}{0.5em} 
\small                      
\bibliographystyle{elsarticle-num}
\bibliography{bibtex-reviews-document-classification, bibtex-reviews-information-fusion,bibtex-information-fusion-document-classification,bibtex-helper}
\endgroup

\end{document}